\pdfoutput=1

\documentclass[11pt]{article}

\usepackage{ACL2023}

\usepackage{times}
\usepackage{latexsym}
\usepackage[T1]{fontenc}

\usepackage[utf8]{inputenc}

\usepackage{microtype}

\usepackage{inconsolata}
\usepackage{comment}
\usepackage{inconsolata}
\usepackage{hyperref}
\usepackage{times}
\usepackage{latexsym}
\usepackage{times}
\usepackage{latexsym}
\usepackage{graphicx}
\usepackage{lipsum}

\usepackage{float}
\usepackage{algorithm}
\usepackage{algorithmic}
\usepackage{placeins}
\usepackage{subfigure}
\usepackage{booktabs}
\usepackage{cleveref}
\usepackage{multirow}
\usepackage{xcolor, soul}
\definecolor{green}{HTML}{C2D8C8}
\definecolor{purp}{HTML}{D2C2D8}
\definecolor{red}{HTML}{F42C14}
\definecolor{dark-red}{HTML}{DE3163}
\definecolor{dark-blue}{HTML}{268EEA}
\definecolor{class}{HTML}{82204A}
\definecolor{struct}{HTML}{558C8C}
\definecolor{retrieval}{HTML}{5C85FF}
\definecolor{qa}{HTML}{EF7B45}
\usepackage{multirow}
\usepackage{rotating}
\definecolor{green}{HTML}{C2D8C8}
\definecolor{purp}{HTML}{D2C2D8}
\definecolor{red}{HTML}{F42C14}
\definecolor{dark-red}{HTML}{FFA500}
\definecolor{dark-blue}{HTML}{268EEA}
\sethlcolor{red}

\newcommand*\samethanks[1][\value{footnote}]{\footnotemark[#1]}
\title{On the Evaluation Practices in Multilingual NLP:\\ Can Machine Translation Offer an Alternative to Human Translations?}
\author{Rochelle Choenni\thanks{$~~$The authors ordered alphabetically with an equal contribution.}$~~^\spadesuit$ \hspace{1em} Sara Rajaee\samethanks$~~^{\diamondsuit}$ \hspace{1em} Christof Monz$^{\diamondsuit}$ \hspace{1em} 
 Ekaterina Shutova$^\spadesuit$ \\
        $~^\spadesuit$ILLC, University of Amsterdam, the Netherlands \\$^\diamondsuit$Language Technology Lab, University of Amsterdam, the Netherlands  \\\texttt{\{r.m.v.k.choenni, s.rajaee, c.monz, e.shutova\}@uva.nl}\\}

\begin{document}
\maketitle
\begin{abstract}
While multilingual language models (MLMs) have been trained on 100+ languages, they are typically only evaluated across a handful of them due to a lack of available test data in most languages. This is particularly problematic when assessing MLM's potential for low-resource and unseen languages. In this paper, we present an analysis of existing evaluation frameworks in multilingual NLP, discuss their limitations, and propose several directions for more robust and reliable evaluation practices. Furthermore, we empirically study to what extent machine translation offers a {reliable alternative to human translation} for large-scale evaluation of MLMs across a wide set of languages. We use a SOTA translation model to translate test data from 4 tasks to 198 languages and use them to evaluate three MLMs. We show that while the selected subsets of high-resource test languages are generally sufficiently representative of a wider range of high-resource languages, we tend to overestimate MLMs' ability on low-resource languages. Finally, we show that simpler baselines can achieve relatively strong performance without having benefited from large-scale multilingual pretraining. \footnote{The code and translated data are available at \url{https://github.com/Sara-Rajaee/mt4multilinguality} }

\end{abstract}

\section{Introduction}

The field of multilingual NLP has seen rapid advances in recent years, both in terms of performance and the coverage of languages. Abundant research on 
multilingual word embeddings, such as MUSE~\citep{lample2018word} and FastText~\citep{bojanowski2017enriching}, quickly gave way to contextualized language models (LMs), where popular models such as BERT~\citep{kenton2019bert} and GPT~\citep{brown2020language}, were extended to the multilingual setting~\citep{ shliazhko2022mgpt}. Due to the LMs reliance on subword tokenization, their vocabularies could naturally be expanded to cover many languages and writing scripts without exploding the model size \cite{sennrich-etal-2016-neural, kudo-richardson-2018-sentencepiece}. Nowadays, we have multilingual language models (MLMs) that have seen 100+ languages during pretraining~\cite{conneau2020unsupervised}. 
In this regard, much effort has been put into enhancing MLMs, i.e., building stronger and larger models with a higher language coverage during pretraining. However, to what extent are our standard practices for the evaluation of 
MLMs accurate and comprehensive? 
In this work, we study the current evaluation practices, including multilingual benchmarks, evaluation setups, and performance interpretation, and discuss their limitations.

In particular, our analysis of popular multilingual evaluation tasks shows that previous works have mostly followed the trends in monolingual NLP in their efforts to scale evaluation to more complex tasks~\citep{liang2020xglue, hu2020xtreme, ruder2021xtreme}. This has left a big gap in the literature on scaling MLM evaluation to cover more test languages~\citep{ponti2020xcopa, srinivasan2021predicting, ahuja2022multi, patankar2022train, hada2023large, dac2023chatgpt, ploeger2024typological}.
While recent MLMs have been shown to perform well on a variety of tasks \cite{xue-etal-2021-mt5, he2023debertav}, current benchmarks still have a limited language coverage, which means that most languages seen during pretraining have never been evaluated on in any task. Moreover, the set of languages that are covered by each task varies considerably. This makes it difficult to systematically compare the abilities of MLMs across languages and tasks. In addition, it has been questioned whether these language subsets are representative for generalization to a wide range of typologically diverse~\citep{ploeger2024typological} and unseen~\citep{ponti2020xcopa} languages. Thus, the applicability of MLMs in creating language technology for the majority of languages is still unexplored.

In this paper, we address this problem by massively scaling the number of test languages to 198 languages, using NLLB ("No Language Left Behind") \cite{nllb2022}, a SOTA machine translation (MT) model. {Given that considerable progress has been made in MT in recent years, we believe that it is worthwhile to reassess the reliability of translated data for evaluation}~\citep{ranathunga2023neural, popel2020transforming}.  
We compare performance of three popular MLMs, i.e., XLM-R {(both base and large versions)}~\citep{conneau2020unsupervised}, BLOOMz~\citep{muennighoff2022crosslingual} and AYA-101~\citep{ustun2024aya}, across 4 tasks and two evaluation frameworks (fine-tuning and zero-shot prompting). 
Our analysis shows that differences in performance on the machine and human-translated data are negligible, 
hence, we believe that MT can offer a reliable {alternative to human translation to} estimate the generalization capabilities of MLMs across a wide range of languages. 

We use the high language coverage in our translated datasets 
to study to what extent the selected language subsets in the original datasets are representative.
We find that language coverage 
tends to be representative of high-resource languages, but for low-resource languages,
current 
selections overestimate MLM ability by up to $7\%$. 
Moreover, we find that 
performance across unseen languages can be surprisingly high compared to a random baseline. 
To put these scores into perspective, we compare MLMs to less powerful baselines and find that simpler models achieve similar results. 
This sheds doubt on whether unseen languages benefit from large-scale multilingual pretraining.

Our contributions can be summarized as follows: (1) We provide a comprehensive overview of the current evaluation practices in multilingual NLP and their limitations and offer a set of recommendations. 
(2) We release our translated test sets and propose a general framework for machine-translating datasets. 
(3) Based on our translated test sets, we show that the performance on 
low-resource languages covered by popular datasets is not sufficiently
representative for a wide range of languages. 
To the best of our knowledge, we are the first to conduct a large-scale evaluation of MLMs using MT. We hope that our contributions 
can aid a more 
comprehensive 
evaluation of MLMs. 

\section{Current Evaluation Practices in Multilingual NLP}



\begin{figure*}[!t]
    \centering
    \scalebox{0.95}{
    \includegraphics[width=\linewidth, height=4.4cm]{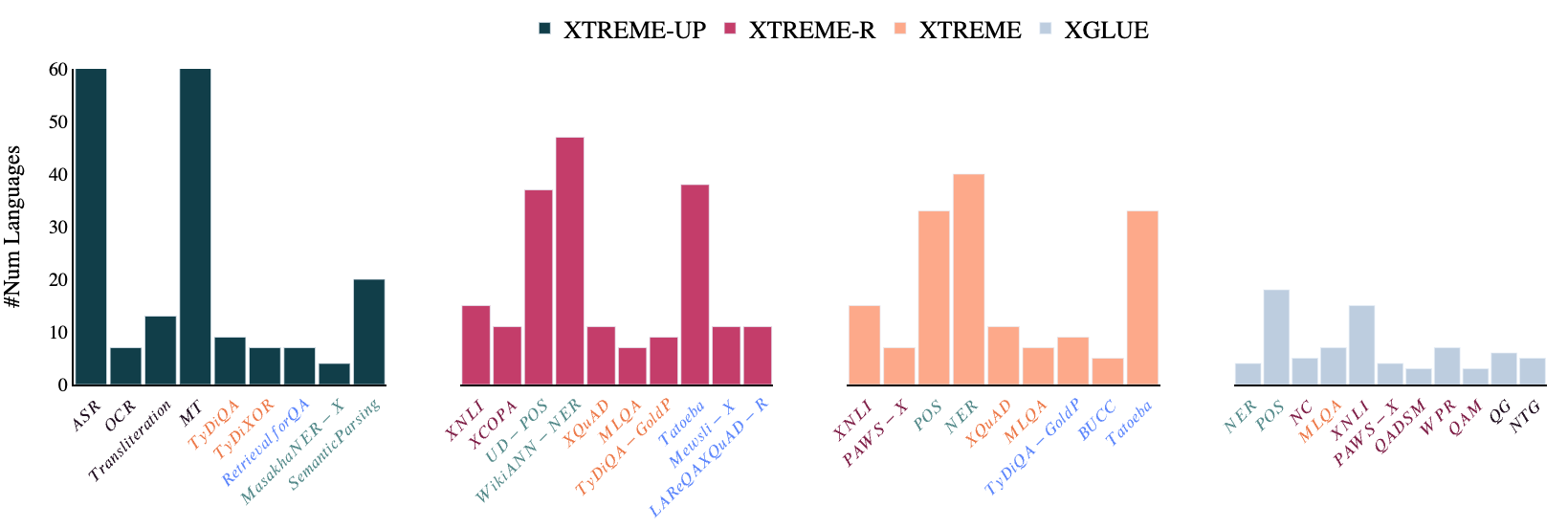}}
    \caption{We report the datasets included in each benchmark along with the number of languages that they cover. The datasets are color-coded by type of task: \textcolor{class}{classification},  \textcolor{retrieval}{retrieval}, \textcolor{qa}{question answering}, or \textcolor{struct}{structured prediction.}}
    \label{fig:overview}
\end{figure*}

\subsection{Multilingual evaluation}\label{sec:evaluation}

Early methods in NLP relied on supervised learning, which is dependent on the availability of manually annotated datasets, 
which are lacking for most languages~\citep{o2016survey}. This made it difficult to extend NLP technology to more languages, and as a result, 
two learning approaches that limit the need for annotated data emerged: language transfer~\citep{ponti2018isomorphic} and joint multilingual learning~\citep{navigli2012joining, ammar2016many}. While the former enables the transfer of models from high to low-resource languages, 
the latter jointly learns from annotated examples in multiple languages to leverage language commonalities. 
Recently, the rise of Transformers~\citep{vaswani2017attention} 
and subword segmentation coupled with multilingual joint learning on the self-supervised masked language modeling task (which does not require annotated data), 
led to the first large-scale MLMs covering 100+ languages. 

Yet, while MLMs circumvent the need for annotated data during pretraining, we still require such data for fine-tuning and evaluation on downstream tasks. Thus, two evaluation approaches have been standardized that do not require large datasets, namely zero-shot and few-shot testing. In the former, the model is fine-tuned on an available dataset for a high-resource language only (typically English) and tested in a target language without having seen any annotated example from it. 
In the few-shot setup, we instead feed the model with data in the test language but only show \emph{few} examples during fine-tuning. Even more recent LLMs like BLOOM~\citep{workshop2022bloom} have been evaluated using zero-shot prompting, where no examples were seen from any language for the task.

While zero-shot and few-shot settings further mitigate the need for annotated data, we still require annotations for testing. As such, researchers started translating English datasets that evaluate the models' ability on a wide variety of semantic and syntactic abilities to multiple languages. However, manual translation is time-consuming, and this has forced the community to focus on a selection of tasks and languages only. Moreover, it has become standard practice to machine translate the training set but use human translation for test sets. To help standardize the evaluation of MLMs, a range of benchmarks have been proposed, e.g., XGLUE~\citep{liang2020xglue}, XTREME~\citep{hu2020xtreme}, XTREME-R~\cite{ruder2021xtreme} and XTREME-UP~\citep{ruder2023xtreme}. These benchmarks encompass a careful selection of challenging multilingual datasets that should provide a comprehensive view of the models' linguistic capabilities. Reporting performance on individual tasks within a benchmark has now been widely accepted as the gold standard for multilingual evaluation.

\subsection{Limitations}\label{sec:limitation}

\subsubsection{Tasks and languages}
\paragraph{Focus on task complexity rather than language coverage}
As more powerful MLMs have been developed, tasks for which a handful of languages achieved high performance have quickly been replaced with tasks that are perceived as more difficult \citep{ruder-etal-2021-xtreme}.
In Figure~\ref{fig:overview}, we depict the evolution of some of the most popular multilingual benchmarks towards more challenging and diverse tasks. 
In particular, we see that the field has mostly moved away from classification tasks and replaced them with question-answering (e.g., XQuAD~\citep{artetxe2020cross} and MLQA~\citep{lewis2019mlqa}) and retrieval (e.g., Tatoeba~\citep{tiedemann2020tatoeba}, BUCC~\cite{zweigenbaum2017overview}) tasks. 
For instance, PAWS-X~\citep{yang2019paws}, a paraphrasing dataset, has been discarded from the XTREME successor, XTREME-R, as it has been claimed that MLMs already beat human performance, see e.g. \citet{tedeschi2023s} for reservations on this. 
However, as the performance on such tasks has in fact never been tested for most languages, we do not believe that 
we can label them as solved just yet.  We argue that in order to discard a task, we need to have reported high performance on at least the languages seen during pretraining. 
\begin{figure}[!t]
    \scalebox{0.85}{
    \includegraphics[width=\linewidth, height=5.5cm]{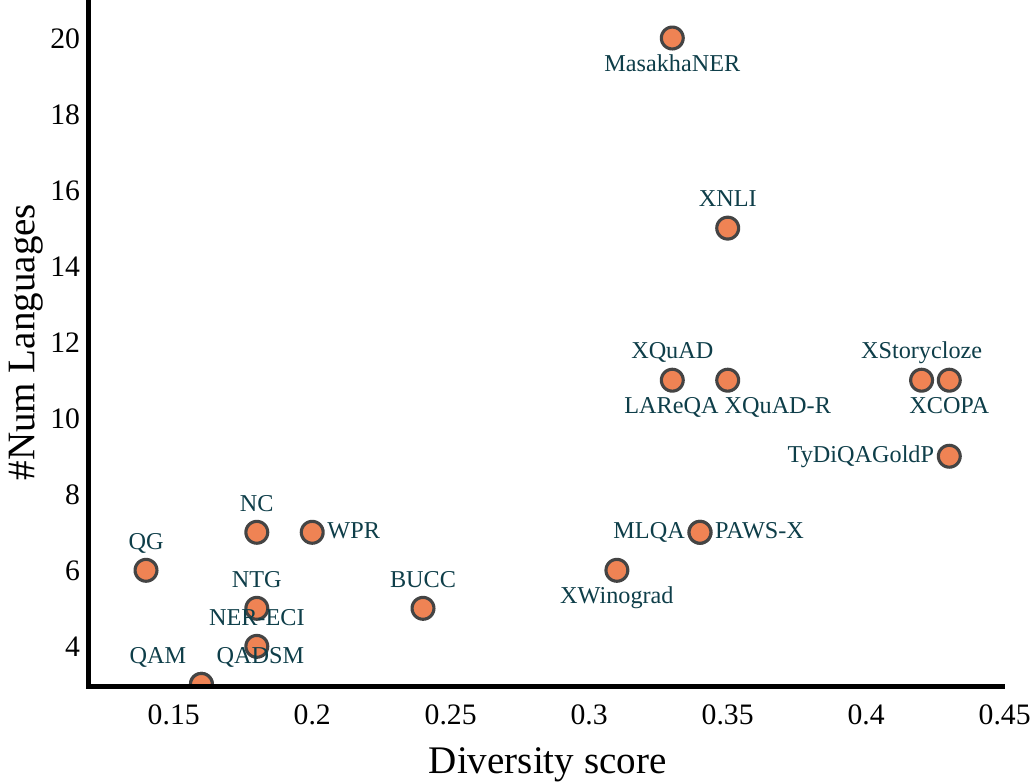}}
    \caption{Number of test languages for each task and the average typological diversity score between them computed as the average cosine similarity between URIEL features of each language pair.}
    \label{fig:typ-div}
\end{figure}
\paragraph{Language coverage is highly skewed to few tasks}
The XGLUE, XTREME, and XTREME-R benchmarks span 19, 40, and 50 languages, respectively. 
However, while the benchmarks span $\pm$20+ languages, it is important to consider that in each benchmark not all tasks cover the same amount of languages. In fact, in Figure~\ref{fig:overview}, we show that the distribution of covered languages is highly skewed towards only a few tasks. For instance, while XTREME-R spans 50 languages, out of the ten proposed tasks, only three, i.e.,\ UD-POS (POS-tagging), WikiANN-NER~\citep{rahimi-etal-2019-massively} (named entity recognition), and Tatoeba (retrieval), cover more than 15 languages. In addition, the subsets of languages covered by each task differ considerably, yet the average performance across tasks is often directly compared.

\paragraph{Language diversity covered by each task varies}
Apart from taking the number of available test languages into account, we should also pay attention to the typological diversity covered by each task. This is particularly needed when we estimate the performance on low-resource languages~\citep{ponti2020xcopa}. 
Typological diversity can be achieved along many different axes, e.g., syntactic features and geographical distance, etc.\ ~\citep{ponti2019modeling}. However, most works underspecify the type of diversity that they consider to obtain a typologically diverse language selection~\citep{ploeger2024typological}. 
In Figure~\ref{fig:typ-div}, we show the differences between the number of test languages of popular tasks and their average typological diversity scores as computed by the cosine similarity between the syntactic URIEL feature vectors from the LANG2VEC library for all test language pairs~\citep{littell2017uriel}. 
While we find that tasks covering more languages also tend to score slightly higher in terms of typological diversity, there are large differences between datasets. 
For instance, while XNLI covers relatively many languages, TyDIQAGolP~\citep{tydiqa} scores much higher on diversity. 

\subsubsection{Dataset construction}
\paragraph{Domain and cross-lingual transfer}
As explained in Section~\ref{sec:evaluation}, multilingual evaluation typically relies on zero-shot or few-shot testing. 
However, in some cases, this means that we are testing the model's ability to perform domain transfer and cross-lingual transfer at the same time~\citep{lai2019bridging}. 
For instance, the Universal Dependency (UD) project~\citep{nivreuniversal} offers consistent treebank annotations across 300+ languages and has become a valuable resource for researchers working on dependency parsing and part-of-speech tagging.
Yet, these treebanks come from vastly different data sources. 
While the commonly used English treebank named `Gum' contains data from 9 different domains (including Wikipedia, spoken language, fiction, etc.), Yoruba and Wolof contain bible texts only. 
Thus, when evaluating performance across languages, this should be taken into consideration, especially as low-resource languages tend to come from different domains more frequently, i.e., spoken language, grammar examples, and bible texts.\footnote{\url{https://universaldependencies.org/}.}

\paragraph{Naturally occurring or parallel data}
While the use of parallel data allows for a controlled evaluation setup that alleviates the aforementioned problem of domain transfer, the use of naturally occurring in-language data also has its benefits. 
In particular, naturally occurring data allows us to test whether the model can handle the nuances and biases that come with different languages~\citep{talat2022you}. 
For instance, \citet{liu2023multilingual} show that proverbs and sayings that occur in different languages are not understood by MLMs, bringing into question whether they can reason in different languages, e.g., `The apple doesn’t fall far from the tree' is understood, yet the model fails on `Bamboo shoots are not far from the clump' a popular saying in Indonesian. 
By limiting MLM evaluation to parallel data---with a mostly Western-centric bias---we can not account for such language characteristics and, therefore, risk overestimating our models' applicability in practice. 

\paragraph{Professional or machine translated data}
While translation always has the disadvantage of adding some possible noise, machine translation also brings the additional problem of the `translationese' effect~\citep{koppel2011translationese}.  
This means that in translation, the sentences in the target language more closely resemble the linguistic patterns from the source language, resulting in phrasings that would not typically occur in the target language. 
As such, it can simplify cross-lingual generalization and cause us to overestimate a model's capabilities in the test language~\citep{zhang2019effect}.

\subsubsection{Interpreting and reporting results}
\paragraph{High- vs. low-resource language categorization differs per model} The amount of data that each MLM has seen during pretraining varies. 
For instance, for XLM-R, which is pre-trained on CommonCrawl, Indonesian is amongst the top 3 most seen languages~\cite{conneau2020unsupervised}, while for mT5 it has the 13th largest dataset~\cite{xue-etal-2021-mt5}, and it has only the 22nd largest dataset in Wikipedia used for training mBERT~\cite{pires-etal-2019-multilingual}. 
In addition, each MLM uses upsampling strategies for their data before pretraining. 
This means that the categorization of high- and low-resource languages is different for each MLM, and we can not compare performance across MLMs using the same resource-based categorization. 
Moreover, the data distribution of pretraining languages is often unclear because (1) developers report this in different formats, e.g., Wikipedia coverage~\citep{wu2020all}, data size in GB~\citep{conneau2020unsupervised}, or percentage towards the full pretraining data \citep{muennighoff2022crosslingual}, (2) the problem of language contamination~\cite{blevins2022language} which means that texts can contain multiple languages, e.g. through quotations or code-switching, making it impossible to know for certain how much data was seen for each language. 
{Lastly, language categorization by data-scarcity is further complicated because many MLMs are instruction-tuned using large datasets. While this can further affect the categorization, these numbers are typically not taken into account~\citep{ustun2024aya}.}

\paragraph{No strong baseline for comparison.}
The performance of MLMs across tasks and languages has usually been compared to previous SOTA models or random baselines.
However, as various works have shown that data artifacts can help the model in obtaining misleadingly high performance~\citep{mccoy2020right}, we question whether random classification is a sufficiently strong baseline~\citep{glavavs2019properly}. 
Especially on low-resource languages, we tend to achieve relatively small improvements over random, thus we believe that it is important to keep evaluating our models against less powerful models to assess to what extent these languages have benefited from large-scale multilingual pretraining.

\section{Methods}
In the next part, we address some of the limitations discussed in Section~\ref{sec:limitation} by machine translating 4 classification datasets into 198 languages. 

\subsection{Tasks and Datasets}

\paragraph{XNLI} The Cross-Lingual Natural Language Inference (XNLI) dataset \citep{conneau2018xnli} contains premise-hypothesis pairs labeled with 
their relationship: `entailment', `neutral', or `contradiction'. The dataset has parallel data in 15 languages. 

\paragraph{PAWS-X} The Cross-Lingual Paraphrase Adversaries from Word
Scrambling (PAWS-X) dataset \citep{yang2019paws} requires the model to determine whether two sentences are paraphrased.
The parallel test data has been provided in 7 languages.

\paragraph{XCOPA}
The Cross-lingual Choice of Plausible Alternatives (XCOPA)~\citep{ponti2020xcopa} evaluates common-sense reasoning abilities in 11 different languages. The samples contain a premise and question paired with two answer choices from which the model can select. 

\paragraph{XStorycloze}
The Cross-lingual Storycloze~\citep{lin2021few} proposes a common-sense reasoning task in 11 languages, in which the model predicts which one of two story endings is the most likely to follow after a given short story. 
\\


\subsection{Language Models}\label{sec:models}

We evaluate the base and large version of XLM-R pre-trained on 100 languages as it is one of the most popular MLMs.
In addition, we report scores from BLOOMz and AYA-101 (AYA).
BLOOMz is trained on 46 languages and AYA on 101, both models are further instruction-tuned on a mixture of prompts in different languages. 
Given that the PAWS-X dataset was included during instruction-tuning, we evaluate BLOOMz and AYA on the held out datasets only. 
Note that BLOOMz and AYA were selected over other MLMs, such as Llama~\citep{touvron2023llama}, as they are the largest publicly available and explicitly MLMs. 

\paragraph{Baselines} As a baseline, we use a vanilla unidirectional LSTM~\citep{hochreiter1997long} with a simple MLP classifier on top. While the model is initialized with the FastText word embeddings, the LSTM is only fine-tuned on the English training data for each task. 
Importantly, the FastText embeddings were pretrained using monolingual data only and later automatically aligned into a shared multilingual vector space~\cite {lample2018word}, see Appendix~\ref{sec:experimental-setups} for more details. 
Thus, using this baseline, we can get a better idea of how much each language has truly benefited from large-scale multilingual pretraining of XLM-R and BLOOMz.

\subsection{Task Evaluation}

\paragraph{Zero-shot testing}
 We fine-tune XLM-R on the entire training set for each task in English. 
We then use our fine-tuned model for zero-shot testing in the test languages. 
We fine-tune our model using the HugginFace Library, 
see Appendix~\ref{sec:experimental-setups} for details.

\paragraph{Zero-shot prompting}
BLOOMz and AYA are instruction-tuned on multiple classification tasks, thus we test these models out-of-the-box in a zero-shot prompting set up. 
This has the benefit that dataset artifacts, which are commonly known to be leveraged during fine-tuning, can not be learned~\cite {mccoy2020right}. 
As BLOOMz and {AYA} fail to predict a third option for XNLI (neutral),  we report results on a binarized version of the task by aggregating sentences with the `neutral' and `contradiction' labels into one class and making the model predict entailment or not. 
Moreover, the instructions are given in English,
see Appendix~\ref{sec:experimental-setups} for the prompts per task. Finally, note that our goal is not to compare performance to XLM-R but rather to test the reliability of machine-translated test sets in two popular evaluation paradigms.

\subsection{Machine Translation}
We employ the NLLB model covering 202 languages for translation \cite{nllb2022}.
To gain a better understanding of the role of the MT system on the translated data quality and, consequently, the performance on downstream tasks, we experiment with two NLLB versions. We choose the distill NLLB with 600M parameters and the 3.3B NLLB model using greedy sampling. 
For each example, we translate each sentence (e.g., \textsc{sentence1} and \textsc{sentence2} in PAWS-X) separately. In Appendix~\ref{sec:lessons}, we provide the lessons learned from our MT experiments.
\begin{table}[t!]
    \centering   
    \setlength{\tabcolsep}{6.5pt}
    \scalebox{0.65}{
    \begin{tabular}{r | c c c c | c}
    \toprule
            & Unseen & Low& Mid & High & Total \\
    \midrule 
     $\%$ of data &  0 & $\textgreater$ 0 and $\textless$ 0.1 &  $\geq$ 0.1 and $\textless$1&  $\geq$1 & \\
    XLM-R & 106 & 30 & 34 & 26 & 196\\
      BLOOMz& 131 & 21 & 7 & 9 & 168\\  
      AYA & 103 & 57 & 25 & 13 & 198\\
    \bottomrule
    \end{tabular} }
    \caption{Number of languages categorized as high, mid, low, and unseen languages when looking at the percentage of seen pretraining data of the respective LMs.}
    \label{tab:categorization}
\end{table}

\begin{table*}[!ht]
    \centering   
    \setlength{\tabcolsep}{2pt}
    \scalebox{0.59}{
    \begin{tabular}{l  c c c c c c c c c c  c c c c c c c c c c c cc cc }
    \toprule
                 & ar & bg & de & el &  es & et &eu & fr& hi & ht& id& it& ja& ko & my& qu& ru& sw &  ta& te& th& tr& ur& vi& zh  \\
                   \midrule 
        & \multicolumn{25}{c}{\textbf{XLM-R}} \\
        \midrule
        
          XCOPA & - & - & - & - & - &  69/73& - & - & - & 50/57 &76/69  &75/73 & - & - & - & 49/59 & - &   66/58& 64/68 & - & 69/66 &71/66& - &68/70& 72/73 \\
            XStoryCloze & 80/80 & - & - & - & 84/84 & - & 79/79 & - & 78/79 & - & 88/87 & - & - & - & 71/70 & - & 84/85 & 75/76 &  -& 76/75 & -& -&-& - & 87/86 \\
            XNLI & 78/78 & 82/82 &82/82 & 81/80 & 83/78 & - & - & 82/83 & 75/80 & - & - & -& - & - & - & -& 79/82 & 70/74 &  -& - & 76/76&77/79& -& 78/81& 79/74 \\
             PAWS-X & - & - & 90/91& - & 91/91 & -& - & 91/90 & - &  -& - & -& 81/84 & 81/83 & - & - & - & - &  -& - & - & - &-& -& 83/82 \\
            \midrule 
        & \multicolumn{25}{c}{\textbf{BLOOMz}} \\
        \midrule
           XCOPA & - & - & - & - & - & 52/52 & - & - & - & N/A & 78/78 & 62/65& - & - & - & 51/52& - & 60/63 &  75/72& - & N/A& 50/50&-& 80/77& 71/67 \\
            XStoryCloze & 88/88 & - & - & - & 91/91 & - & 84/78 & - & 85/84 & - & 91/90 & - & - & - & 54/52 & - & 73/73 & 79/79 &  -& 74/73 & -& -&-& - & 70/70 \\
            B-NLI & 71/72 & 66/68 & 69/68 & 65/66 & 73/74 & - & - & 72/73 & 70/72 & - & - & -& - & - & - & -& 69/70 & 70/71 &  -& - &  N/A&68/70&-& 72/73& 74/72 \\
            \midrule
        & \multicolumn{25}{c}{\textbf{AYA}} \\
        \midrule
           XCOPA & - & - & - & - & - & 87/84 & - & - & - & 82/83 & 87/87 & 88/88& - & - & - & 56/56& - & 79/83 &  86/83& - & 84/82& 86/85&-& 85/84& 86/84 \\
            XStoryCloze & 95/92 & - & - & - & 94/94 & - & 83/75 & - & 93/91 & - & 91/87 & - & - & - & 94/86 & - & 90/82 & 93/89 &  -& 93/88 & -& -&-& - & 95/ 90\\
            B-NLI & 78/79 & 79/79 & 78/78  & 78/78   & 79/80  & - & - & 79/80  & 75/75  & - & - & -& - & - & - & -& 79/79  & 74/75  &  -& - &  79/79 &79/79 &-& 76/77 & 77/77  \\
\bottomrule
\end{tabular}}
    \caption{{The ($\%$) accuracy of the models on the human translated (original)/our machine translated datasets.}} 
    \label{tab:translation-quality}
\end{table*}

\begin{table*}[t!]
    \centering   
    \setlength{\tabcolsep}{6.5pt}
    \scalebox{0.7}{
    \begin{tabular}{r | c c c c |c c c |c c c  }

    \toprule
    \multicolumn{1}{c}{} & \multicolumn{4}{c}{\textbf{XLM-R}} & \multicolumn{3}{c}{\textbf{BLOOMz}} & \multicolumn{3}{c}
    {\textbf{AYA}}  \\
    \midrule
    
            & XCOPA & XNLI & PAWS-X & XStoryCloze  & XCOPA & XNLI & XStoryCloze    & XCOPA & XNLI & XStoryCloze\\
    \midrule
    Pearson corr. & 95.3 & 69.3 & 93.3 & 98.0 &
    85.0 & 97.8 & 98.7 & 98.0 & 95.3 & 90.7 \\
    Spearman rank corr. & 77.5 & 82.4 & 82.6 & 98.8 &
    89.0 & 95.1 & 97.3 & 81.8 & 92.5 & 77.0\\

\bottomrule
\end{tabular}}
    \caption{ 
     Pearson and Spearman rank correlation between human- (original data) and machine-translated data (ours). } 
    \label{tab:spearman-correlation}
\end{table*}

\paragraph{Evaluation metric}
As an evaluation metric for testing our machine translation quality, we use the chrF++ metric~\cite{popovic-2017-chrf}. chrF++ calculates the character and word n-gram overlap between the machine and human reference translation.
It is a tokenization-independent metric aligning better with human judgments for morphologically-rich languages compared to BLEU \cite{tan-etal-2015-awkward,  kocmi-etal-2021-ship, briakou-etal-2023-searching}. 
For the evaluation, we use the FLORES-200 dataset~\cite{nllb2022}, which includes human-translated data for 200 languages, and select 100 sentences from each language.

\subsection{Selection of Test Languages}
For our selection of test languages, there are two constraining factors: (1) the language has to be covered by the NLLB-200 translation model and FLORES-200 dataset, and (2) the script of the language needs to have been seen during the pretraining of the model. 
We then separately filter out the test languages by unseen scripts for each model. 
This leaves us with 196, 168, and 198 test languages for XLM-R, BLOOMz, and AYA, respectively, see Appendix~\ref{sec:languagcoverage} for the complete lists. 
Note that the number of compatible languages is lower for BLOOMz as it has seen fewer writing scripts during pretraining.
Moreover, we categorize the test languages for each model separately based on the percentage of total data that they contributed during pretraining. 
In Table~\ref{tab:categorization}, we report the percentage thresholds used for our categorization and the resulting number of test languages for each category and model.
As the pretraining data coverage is reported in numbers of GB for XLM-R, we convert these scores to percentages of the full pretraining data. 
For BLOOMz, we use the reported language distribution numbers \footnote{\url{https://huggingface.co/bigscience/bloom}}, and for AYA, we consider mT5 pretraining data distribution as it is utilized as the base model for AYA.

\section{Results on Machine Translation Quality}
MT systems are known to have problems~\citep{artetxe2020cross} such as hallucination~\citep{wang2020exposure}, or translationese output, but recent models have shown considerable advances, especially on low-resource languages~\citep{ranathunga2023neural}. To ensure that this is indeed the case, we now validate the high quality of our translations using three different analyses.



\paragraph{ChrF++ scores} In Table~\ref{tab:translation-scores}, we report the average chrF++ scores for translations obtained with the NLLB-Distil and NLLB-3.3B models on the dev set of FLORES-200 dataset including human translation data in 200 languages~\cite{nllb2022}. 
We categorize languages by XLM-R resource ranking (high, mid, low, and unseen) and observe that while the quality for high-resource languages is comparable, for low-resource languages, NLLB-3.3B tends to perform substantially better. 
Thus, in all further experiments, we use NLLB-3.3B for translation. Moreover, we confirm that our scores for all languages are among the best performances of the SOTA multilingual MT systems \cite{bapna2022building, nllb2022}.

\paragraph{Comparison to professional translations} To test to what extent machine-translated data causes differences in performance compared to the professionally translated data, we run an evaluation on both subsets.
In Table~\ref{tab:translation-quality}, we report the performance obtained using machine and professionally translated data, respectively.\footnote{Results of XLM-R base are reported in the appendix. }
We find that 
on average, the difference is only 2.6\% in accuracy scores. 
Moreover, we see that the same trend holds for all models.
Thus, in both evaluation setups, the effect of MT quality is relatively small, making MT an acceptable alternative to human translation. 
Finally, in some cases, we find that MT slightly increases performance. We suspect that this is due to the `translationese' effect explained in Section~\ref{sec:limitation}.

\paragraph{Correlation between translation quality and model performance}
Finally, to test how reliable machine translation is in the scope of our study, we report the average Pearson and Spearman rank correlation between the models' performance using human-translated and machine-translated data, see Table~\ref{tab:spearman-correlation}. The high correlation shows that 
the rank order of the models' performance across different languages using human and machine-translated data is almost the same. 
Moreover, we compute the correlation between the MT quality measured by chrF++ and the model performance on human and machine-translated data; see Appendix \ref{sec:corr} for the numerical results. The results show that there is no meaningful correlation between the performance on the human-translated data and MT quality, and this pattern holds for our machine-translated data.  
\\
\noindent {Based on the results from all three analyses, we conclude that the MT quality is sufficient and results in model performance close to when using human translation.}


\section{Large-scale Evaluation Results}
\begin{figure*}[!t]
    \centering
    \scalebox{0.7}{
     \includegraphics[width=0.49\linewidth]{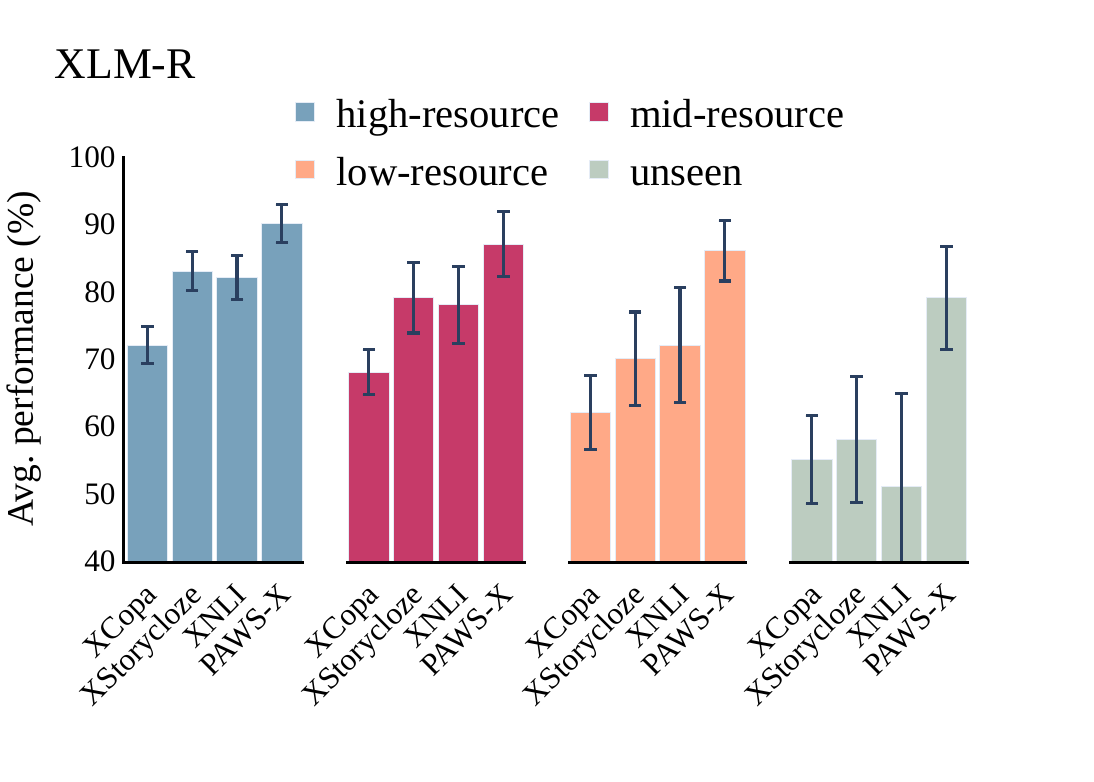}
     \includegraphics[width=0.49\linewidth]{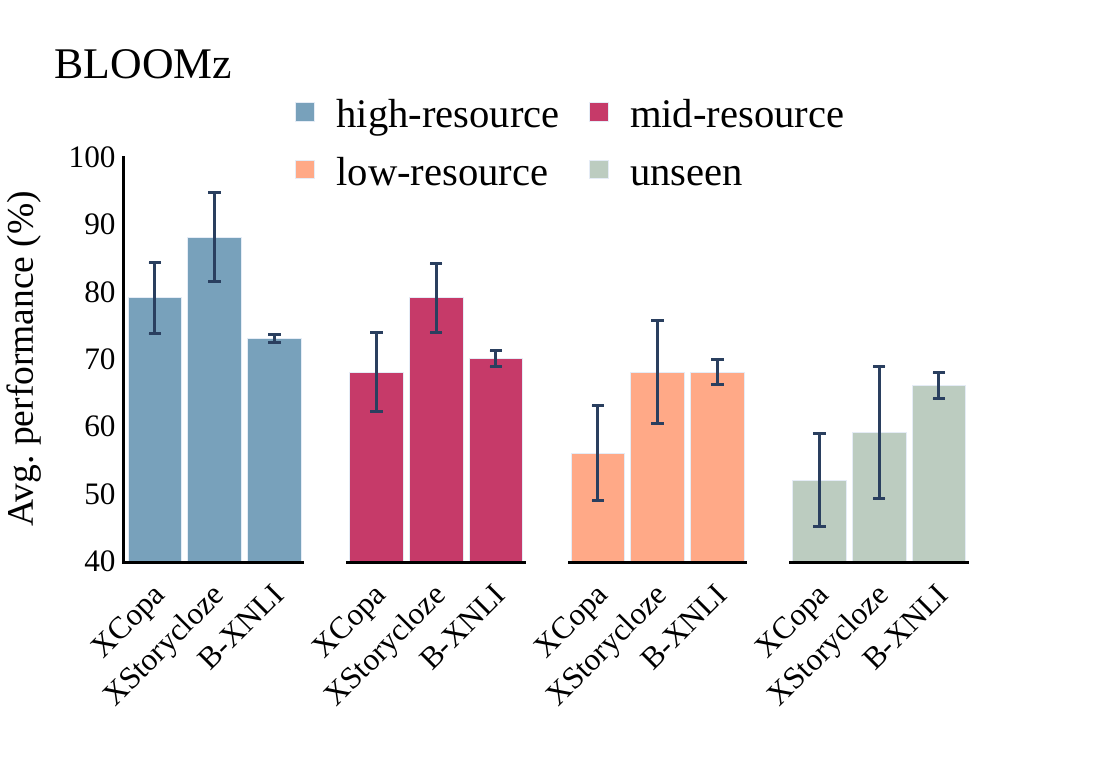}
       \includegraphics[width=0.49\linewidth]{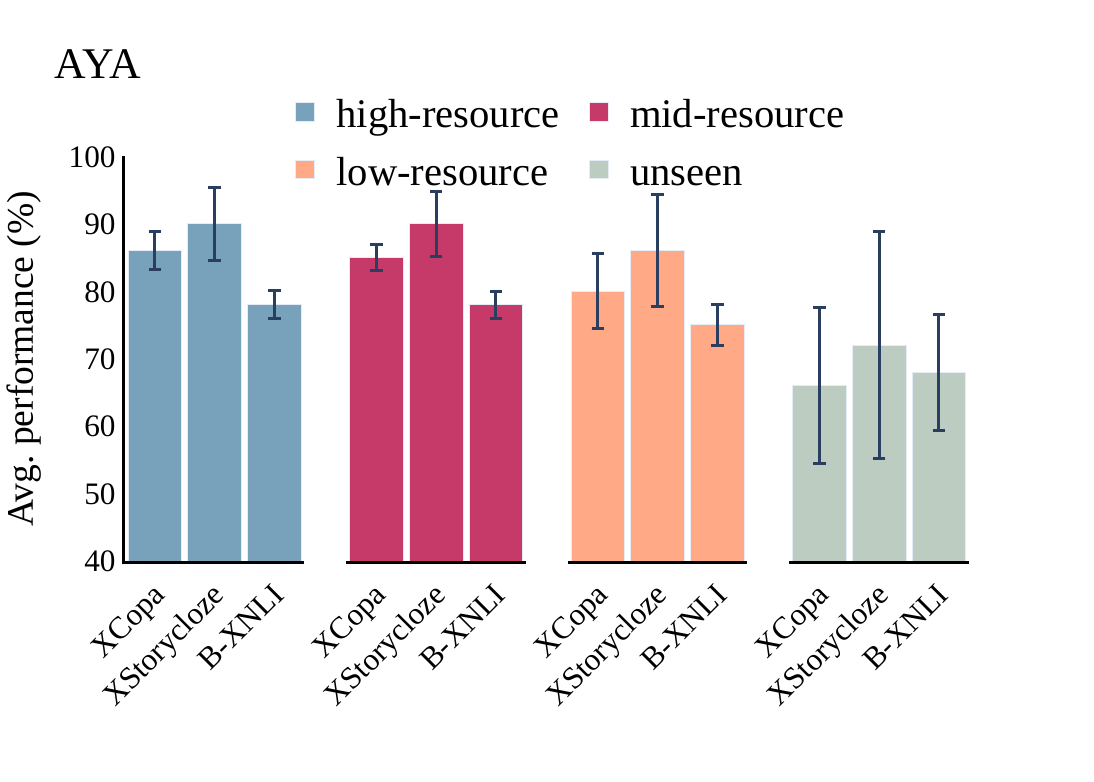}}
     \vspace{-0.5cm}
    \caption{Average performance across test languages in a zero-shot fine-tuning setup for XLM-R and in a zero-shot prompting using BLOOMz. Results are categorized per task and data coverage during pretraining as reported in Table~\ref{tab:categorization}. Results across models are not directly comparable as their language categorizations differ. }
    \label{fig:bloomz-xlmr}
\end{figure*}

Having confirmed that our translated test sets have a reliable quality, we now move on to analyze how MLMs perform on them.
{In Figure~\ref{fig:bloomz-xlmr}, we summarize performance from XLM-R, BLOOMz, and AYA across 196, 168, and 198 languages, categorized by their data-scarcity.}
We find that the average performance for all models is similar for high- and mid-resource languages. 
Yet, while still above the random baseline, there is a notable drop in performance for low-resource and unseen languages.
Moreover, we find that their standard deviations are larger than high- and mid-resource ones. 
This shows performance across low-resource languages varies a lot, making the average score less reliable. 
Still, performance on unseen languages is relatively high; for XNLI and PAWS-X, on average, we obtain 18$\%$ and 29$\%$ above random performance. 

\subsection{Representativeness of Selected Subsets of Languages}
As each dataset contains a distinct selection of languages for testing, we study to what extent each of them provides a reliable estimate for how MLM performance will generalize to more languages. 
While we do not cover all the world's languages, we compare the averages between the languages covered by the original datasets and those covered by our much larger translated datasets. 
To this end, we split the languages from the original datasets based on our resource categorization reported in Table~\ref{tab:categorization}, and report the average performance for each category in Table~\ref{tab:representativeness-peformance}. 
Importantly, all performance scores are computed on the translated data. 
From the results, we observe that for high and, to some extent, mid-resource languages, average performance on both language selections is similar, making the language coverage from the original datasets sufficiently representative.
Yet, for low-resource languages, we find a notable difference, which suggests that the datasets' language coverage is not representative of a wider range of low-resource languages. 
Specifically, across all tasks, we tend to overestimate performance, which can go up to 4.7$\%$ and 8.7$\%$ accuracy points (for XCOPA).

\begin{table}[t!]
    \centering   
    \setlength{\tabcolsep}{6.5pt}
    \scalebox{0.65}{
    \begin{tabular}{r | c c c |c  }

    \toprule
            & High & Mid & Low & Ave. \\
    \midrule 
     \multicolumn{5}{c}{\textbf{XLM-R}} \\
    \midrule
    PAWS-X
    & 89.5 / 89.6
    & ~~~-~~ / 87.3
    & ~~~-~~ / 85.6
    & 89.5 / 85.3
    \\
    XNLI
    & 82.2 / 81.5
    & 79.5 / 78.2
    & 74.4 / 71.9
    & 80.5 / 72.4
    \\
    XCOPA
    & 71.6 / 71.8
    & 69.6 / 68.2
    & 66.4 / 61.7
    & 70.3 / 69.2
    \\
    XStoryCloze
    & 85.5 / 83.1
    & 77.2 / 78.6
    & 74.1 / 69.9
    & 78.9 / 77.2
    \\
    \midrule 
     \multicolumn{5}{c}{\textbf{BLOOMz}} \\
     \midrule
     B-NLI
    & 72.8 / 73.0
    & 70.8 / 70.2
    & 70.9 / 68.3
    & 72.2 / 69.8
    \\
    XCOPA
    & 76.9 / 79.0
    & 71.6 / 67.5
    & 63.0 / 54.3
    & 73.7 / 62.3
    \\
    XStoryCloze
    & 86.2 / 87.9
    & 81.1 / 72.0
    & 79.4 / 64.5
    & 82.8 / 71.6
    \\
     \midrule 
     \multicolumn{5}{c}{\textbf{AYA}} \\
     \midrule
     B-NLI
    & 78.4/77.8
    & 77.7/77.6
    & 73.5/75.5
    & 77.4/76.4
    \\
    XCOPA
       & 83.65/86.3
    & 84.8/84.6
    & 82.8/79.8
    & 83.7/82.0
    \\
    XStoryCloze
    & 85.8/89.7
    & 91.0/89.6
    & 85.5/86.4
    & 87.0/87.7
    \\

\bottomrule
\end{tabular}}

    \caption{ 
    The average performance of high, mid, and low-resource languages covered by the original dataset/the languages covered by our machine-translated datasets. 
    All results are computed on the machine-translated data.} 
    \label{tab:representativeness-peformance}
\end{table}

\begin{table}[t!]
    \centering   
    \setlength{\tabcolsep}{6.5pt}
    \scalebox{0.65}{
    \begin{tabular}{r | c c c c c c c c }
    \toprule
            & en & ar & bg & de & el & es & fr & hu \\
           
    \midrule
     MLP  & 54.6 & 53.8 & 56.6 & 55.2 & 54.4 & 55.3 & 56.4 & 53.8 \\
     LSTM & 74.5 & 68.1 & 72.9 & 73.4 & 68.7 & 77.7 & 78.9 & 72.5  \\

     \midrule
       & id & mk & no & ro & ru & sk & tr & vi \\
       \midrule
       MLP  & 55.0 & 56.6 & 55.8 & 56.8 & 56.1 & 55.7 & 54.9 & 54.7 \\
     LSTM & 73.1 & 72.2 & 74.7 & 75.0 & 74.3 & 71.7 & 68.4 & 85.1 \\
     
    \bottomrule
    \end{tabular} }
    \caption{The performance of the baselines on XNLI. }
    \label{tab:baseline}
\end{table}

\subsection{Baseline Performance}
While more powerful MLMs have been proposed in recent years, random baselines are still commonly used to interpret if performance improvements are meaningful.
We argue that having a stronger yet simpler baseline can better put performance improvements into perspective. 
Thus, 
we use an LSTM and the FastText embeddings with an MLP on top for XNLI and PAWS-X, as we see unexpectedly high performance for unseen languages for these two tasks. 
Table~\ref{tab:baseline} summarizes our results across 16 languages, see Appendix~\ref{app:baseline-paws} for PAWS-X. 
As can be seen, these simple models can impressively achieve high performance across different languages. 
Especially among low-resource languages like Turkish, there is no notable difference between these models and our advanced MLMs. We conclude that even though we have used much more computation and data for pretraining MLMs, their performance has not improved proportionally to this on mid and low-resource languages.






\section{Conclusions and Recommendations}

{We demonstrate that machine translation provides a valid alternative to human translation of test sets.} Employing our large-scale translated test sets, we show that the subsets of languages selected for each dataset tend to give a misleading estimate of MLM performance on low-resource languages. In addition, we show that random performance is a relatively weak baseline to determine meaningful performance improvement. Taking into consideration some of the limitations discussed in this work, we provide a set of practical recommendations for future work: 
(1) \emph{Reconsidering MT for evaluation}: While MT has its shortcomings, we show that MT quality is sufficient to accurately estimate performance across a wide range of languages. 
(2) \emph{Reconsidering simpler baselines}: 
We suggest reconsidering simpler neural networks to test the extent to which an MLM's large-scale pretraining has benefited performance. (3) \emph{Paying attention to the categorization of high- and low-resource languages}: To compare MLMs' performance, it is important to pay attention to the different categorizations of high- and low-resource languages for each model separately.
(4) \emph{Keeping domain sources consistent between languages}: For a fair comparison across languages, it is crucial to make sure that test data come from similar domain sources. 
We believe that these practices improve evaluation and interpretation of performance in multilingual NLP.

\section*{Limitations}
 Our machine translation alternative might only be applicable to the evaluation of classification tasks. For more complicated tasks, translation noise is more likely to cause a bigger performance gap compared to human-translated test sets.
 In Appendix~\ref{sec:lessons}, we provide a set of lessons learned from our MT experiments and found that for automatically translating questions-answering datasets, for instance, we face more challenges than when simply translating classification tasks. 
 Besides, when using machine translation, different cultural information is not transferred from one language to another. This could make it more difficult to generalize across languages. As such, we are probably still overestimating the models' ability to perform cross-lingual transfer when testing on parallel data.
In addition, due to computational limitations, we only use the NLLB 3.3B model. However, as we saw in Table \ref{tab:translation-scores}, a bigger version of the MT model can enhance the translation quality, especially for low-resource languages. This suggests that using even larger models could have made the performance even more reliable.

Moreover, in this paper, we used the percentage of data that each language contributes to the total amount of pretraining data as a measure of data coverage. Based on this measure, we set thresholds to distinguish high, mid, and low resource languages. However, this value is relative to the total amount of seen pretraining data, which can differ per LM. One could argue that these threshold values should instead be based on the absolute amount of data seen during pretraining. 
Future work should focus on studying whether the categorization of data coverage should be made based on absolute or relative numbers.



\bibliography{anthology,custom}
\bibliographystyle{acl_natbib}

\appendix
\clearpage

\section{Experimental setups}
\label{sec:experimental-setups}

For the implementation of all models, we rely on the HuggingFace Library~\citep{wolf2019huggingface}. XLM-R large, BLOOMz, and AYA-101 (AYA) have ~330M, 7.1B, and 13B parameters, respectively.  Moreover, we have run all the BLOOMz and AYA experiments on an NVIDIA A100-SXM4 GPU with 40GB memory, and a single NVIDIA A6000 has been used for the MT and XLM-R experiments with 48GB memory.

\paragraph{XLM-R fine-tuning details}
For the NLI task, we have fine-tuned XLM-R with a learning rate of 2$e$-5, AdamW optimizer, and a batch size of 32 for $3$ epochs. For the PAWS-X task, we have considered a learning rate of 2$e$-6, batch size 16 with a warm-up ratio of $0.01$ for 3 epochs. For XCOPA and XStoryCloze tasks, first, we train the model on the training set of Social IQa \cite{sap-etal-2019-social} and then fine-tune it on the training set of XCOPA dataset \cite{gordon-etal-2012-semeval}. We have selected a learning rate of  3$e$-6, batch size of $16$ for SIQa and $8$ for XCOPA, a warm-up ratio of $0.1$, and fine-tune the model for $3$ epochs on each dataset.

\paragraph{BLOOMz and AYA zero-shot prompts}
For zero-shot prompting, we constructed the following prompts for XNLI, XCOPA and XStorycloze respectively:

\begin{quote}
    Premise: \texttt{<premise>}\\
    Hypothesis: \texttt{<hypothesis>}\\
    Does the premise entail the hypothesis?
    Pick between yes or no.
\end{quote}
-----------------------------------------------------
\begin{quote}
    Premise: \texttt{<premise>}\\
    Option A:  \texttt{<choice1>}\\
    Option B:  \texttt{<choice2>}\\
    Based on the premise, which \texttt{<cause/effect>} is more likely? Pick between options A and B.\\
    Answer: 
\end{quote}
-----------------------------------------------------
\begin{quote}
    Consider the following story:\\
    \texttt{<story>}\\
    Which ending to the story is most likely?\\
    Pick between options A and B:\\
    A: \texttt{<story\_ending1>}\\
    B: \texttt{<story\_ending2>}\\
    Answer: 
\end{quote}

\paragraph{Baseline fine-tuning details}
We have trained the LSTM with a hidden size of 300, a learning rate of 4$e$-4, and a batch size of 32 for 64 epochs with an early stopping strategy. The MLP classifier has a 2-layer architecture (same as the XLM-R fine-tuning setup) with the $tanh$ function as the non-linearity. 
We have used the MLP model with the same hyperparameters for the MLP experiments.
For all the evaluations, we have utilized our translated datasets.

\begin{table}[t!]
    \centering   
    \setlength{\tabcolsep}{6.5pt}
    \scalebox{0.65}{
    \begin{tabular}{r | c c c c |c c }

    \toprule
           \textbf{metric}   & High & Mid & Low & Unseen & Ave. & Med. \\
    \midrule 
    \multicolumn{7}{c}{\textbf{NLLB-Distil}}\\
    \midrule
    chrF++
    & 49.34
    & 46.92
    & 43.22
    & 37.12
    & 44.15
    & 45.07
    \\
    \midrule 
    \multicolumn{7}{c}{\textbf{NLLB-3.3B}}\\
    \midrule
    chrF++
    & 52.5
    & 50.8
    & 46.1
    & 40.3
    & 44.5
    & 45.5
    \\

\bottomrule
\end{tabular}}

    \caption{The quality of the MT system across high, mid, and low-resource languages using chrF++ based on the XLM-R model's categorization. }
    \label{tab:translation-scores}
\end{table}

\section{Correlation between Chrf++ scores and translations}\label{sec:corr}
\begin{table*}[t!]
    \centering   
    \setlength{\tabcolsep}{6.5pt}
    \scalebox{0.65}{
    \begin{tabular}{r | c c c c |c c c |c c c  }

    \toprule
    \multicolumn{1}{c}{} & \multicolumn{4}{c}{\textbf{XLM-R}} & \multicolumn{3}{c}{\textbf{BLOOMz}} & \multicolumn{3}{c}
    {\textbf{AYA}}  \\
    \midrule
    
            & XCOPA & XNLI & PAWS-X & XStoryCloze  & XCOPA & XNLI & XStoryCloze    & XCOPA & XNLI & XStoryCloze\\
    \midrule
    chrF++ vs. Human translated  & 29.6 & 24.4 & 65.7 & 16.4 & 
    5.7 & 5.7 & 68.1 & 22.7 & 8.1 & -34.6\\
    chrF++ vs. Machine translated & 22.6 & 48.1 & 82.7 & 22.4 & 18.3 &
    14.7 & 75.8 & 55.5 & 26.1 & 22.4 \\
\bottomrule
\end{tabular}}
    \caption{ 
    Average Spearman rank correlation between chrf++ scores and human- (original data) and machine-translated data (ours). } 
    \label{tab:spearman-correlation}
\end{table*}

\begin{table*}[t!]
    \centering   
    \setlength{\tabcolsep}{6.5pt}
    \scalebox{0.65}{
    \begin{tabular}{r | c c c c |c c c |c c c  }

    \toprule
    \multicolumn{1}{c}{} & \multicolumn{4}{c}{\textbf{XLM-R}} & \multicolumn{3}{c}{\textbf{BLOOMz}} & \multicolumn{3}{c}
    {\textbf{AYA}}  \\
    \midrule
    
            & XCOPA & XNLI & PAWS-X & XStoryCloze  & XCOPA & XNLI & XStoryCloze    & XCOPA & XNLI & XStoryCloze\\
    \midrule
    chrF++ vs. Human translated  & 27.4 & 10.5 & 89.3 & 3.3 & -10.4 &
    -10.4 & 64.5 & 41.0 & 8.2 & -18.5\\
    chrF++ vs. Machine translated & 30.3 & 66.2 & 96.5 & 8.8 & 14.5 & 2.9 & 66.7 & 52.1 & 16.3 & 11.3 \\

\bottomrule
\end{tabular}}

    \caption{ 
    Average Pearson correlation between chrf++ scores and human- (original data) and machine-translated data (ours). } 
    \label{tab:pearson-correlation}
\end{table*}

\section{Lessons learned for MT}\label{sec:lessons}
We now share a few lessons learned from MT using NLLB to facilitate the translation of new datasets in future work:
\begin{itemize}
    \item NLLB tends to skip sentences when translating paragraphs. Thus, it is important to translate the sentences one by one. 
    \item NLLB has difficulty translating short phrases/names such as names, dates, locations, etc., because it tends to hallucinate additional content. This makes it challenging to translate the answers from QA datasets such as XQuAD. 
    \item NLLB inconsistently chooses to code-switch to the target language. For instance, when translating the sentence `Sara is asleep', it can choose to translate it either to `Sara ? farsi' or `fully farsi'. This can be particularly challenging for retrieval datasets where the answer does not tend to fully match the context. 
    \item While translation quality tends to be similar for different NLLB model sizes, at least the 3.3B version should be used when translating to languages that were low-resource, considering NLLB's pretraining data. 
\end{itemize}

\section{Baselines for PAWS-X}
\begin{table}[h!]
    \centering   
    \setlength{\tabcolsep}{6.5pt}
    \scalebox{0.65}{
    \begin{tabular}{r | c c c c c c c c }
    \toprule
            & en & ar & bg & de & el & es & fr & hu \\
    \midrule
     MLP  & 57.6 & 57.1 & 57.8 & 58.2 & 57.5 & 58.3 & 57.8 & 58.9  \\
     LSTM & 58.5 & 57.3 & 57.9 & 60.3 & 58.8 & 58.8 & 57.8 & 59.8 \\
     \midrule
       & id & mk & no & ro & ru & sk & tr & vi \\
       \midrule
      MLP  & 59.0 & 57.4 & 58.6 & 59.4 & 57.5 & 58.5 & 58.6 & 56.1 \\
     LSTM  & 60.3 & 58.9 & 60.4 & 60.6 & 57.6 & 59.3 & 60.4 & 57.1\\
     
    \bottomrule
    \end{tabular} }
    \caption{The performance of the baselines on PAWS-X. }
    \label{tab:baseline-paws}
\end{table}

\clearpage
\onecolumn

\section{Full results XLM-R Large}
\label{app:baseline-paws}
\begin{figure}[H]
    \centering
    \scalebox{1}{
    \includegraphics[width=\linewidth]{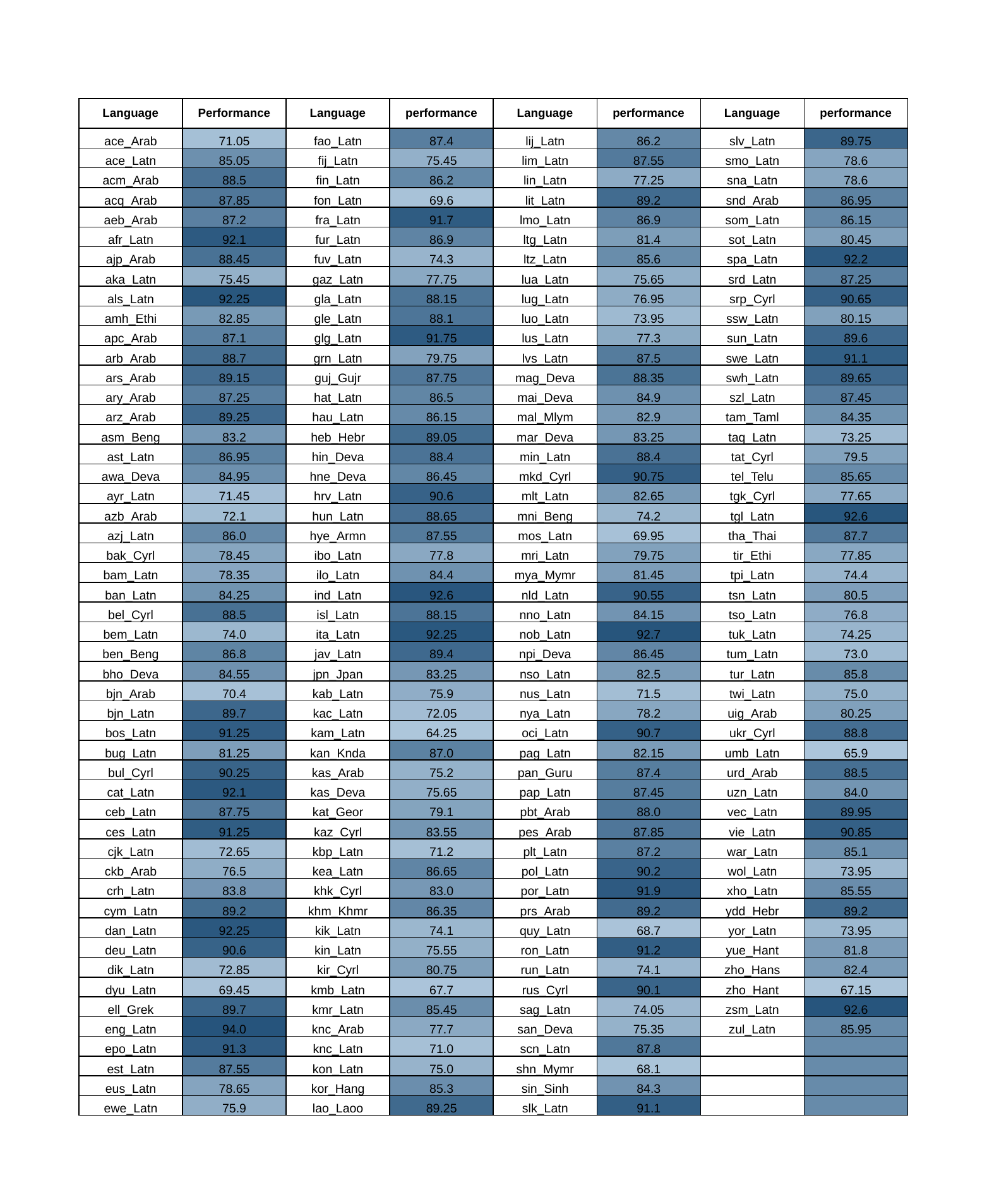}}
    \caption{The accuracy score of XLM-R model on PAWS-X task across 196 languages.}
    \label{fig:xlmr-paws-detailed-results}
\end{figure}

\begin{figure}[H]
    \centering
    \scalebox{1}{
    \includegraphics[width=\linewidth]{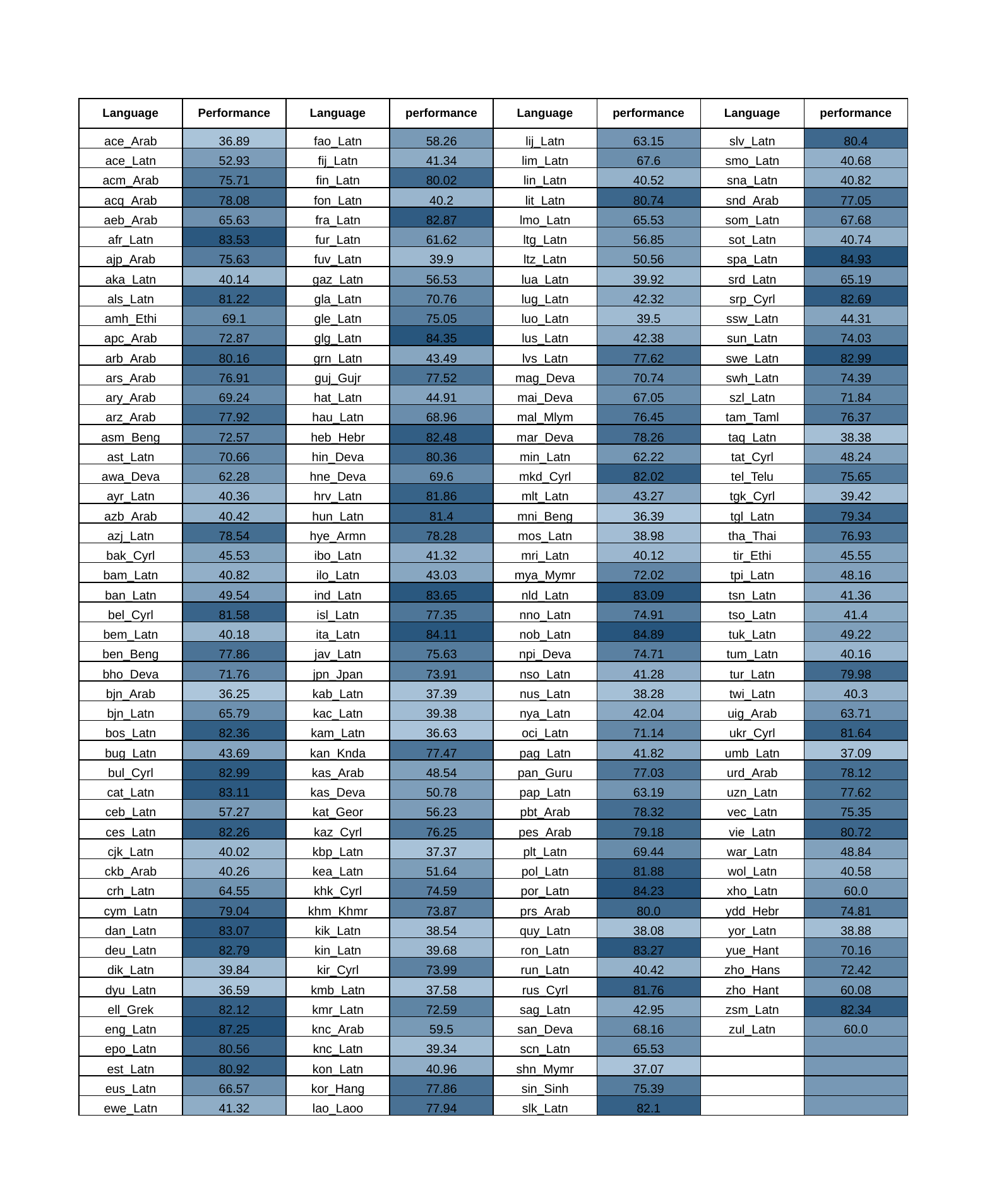}}
    \caption{The accuracy score of XLM-R model on XNLI task across 196 languages.}
    \label{fig:xlmr-xnli-detailed-results}
\end{figure}

\begin{figure}[H]
    \centering
    \scalebox{1}{
    \includegraphics[width=\linewidth]{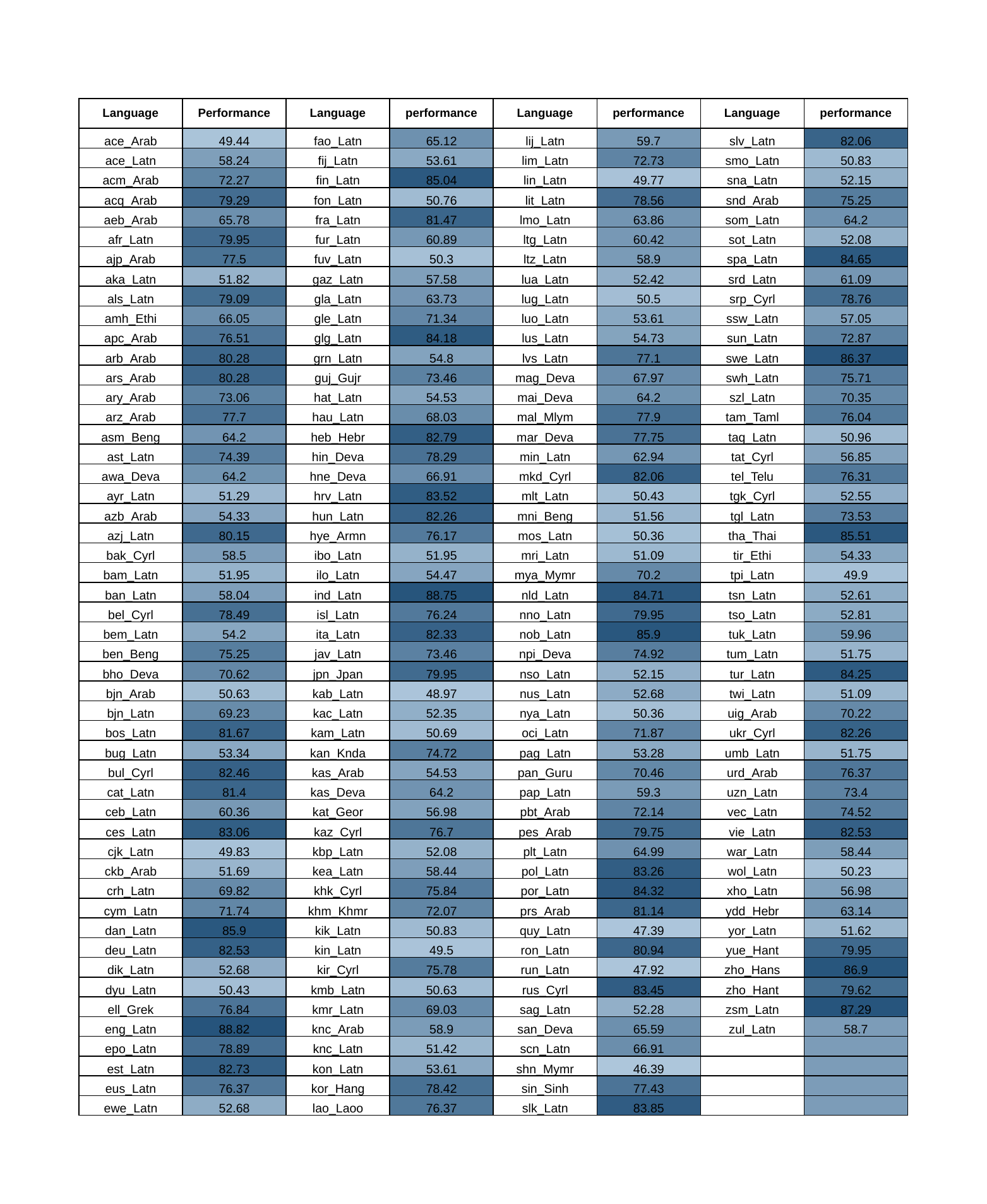}}
    \caption{The accuracy score of XLM-R model on XStoryCloze task across 196 languages.}
    \label{fig:xlmr-story-detailed-results}
\end{figure}

\begin{figure}[H]
    \centering
    \scalebox{1}{
    \includegraphics[width=\linewidth]{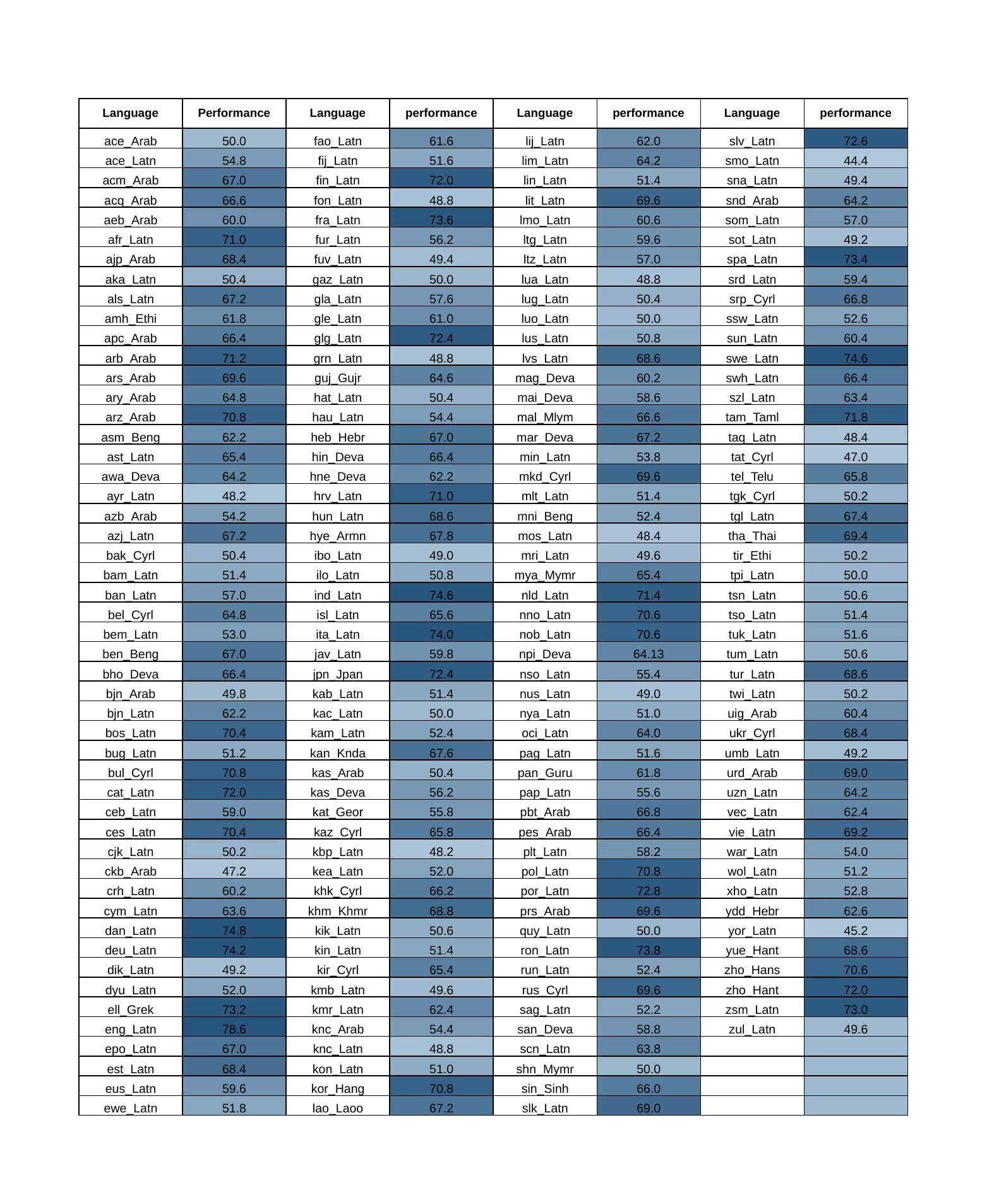}}
    \caption{The accuracy score of XLM-R model on XCOPA task across 196 languages.}
    \label{fig:xlmr-copa-detailed-results}
\end{figure}

\section{Full results BLOOMz}
\begin{figure}[H]
    \centering
    \scalebox{1}{
    \includegraphics[width=\linewidth]{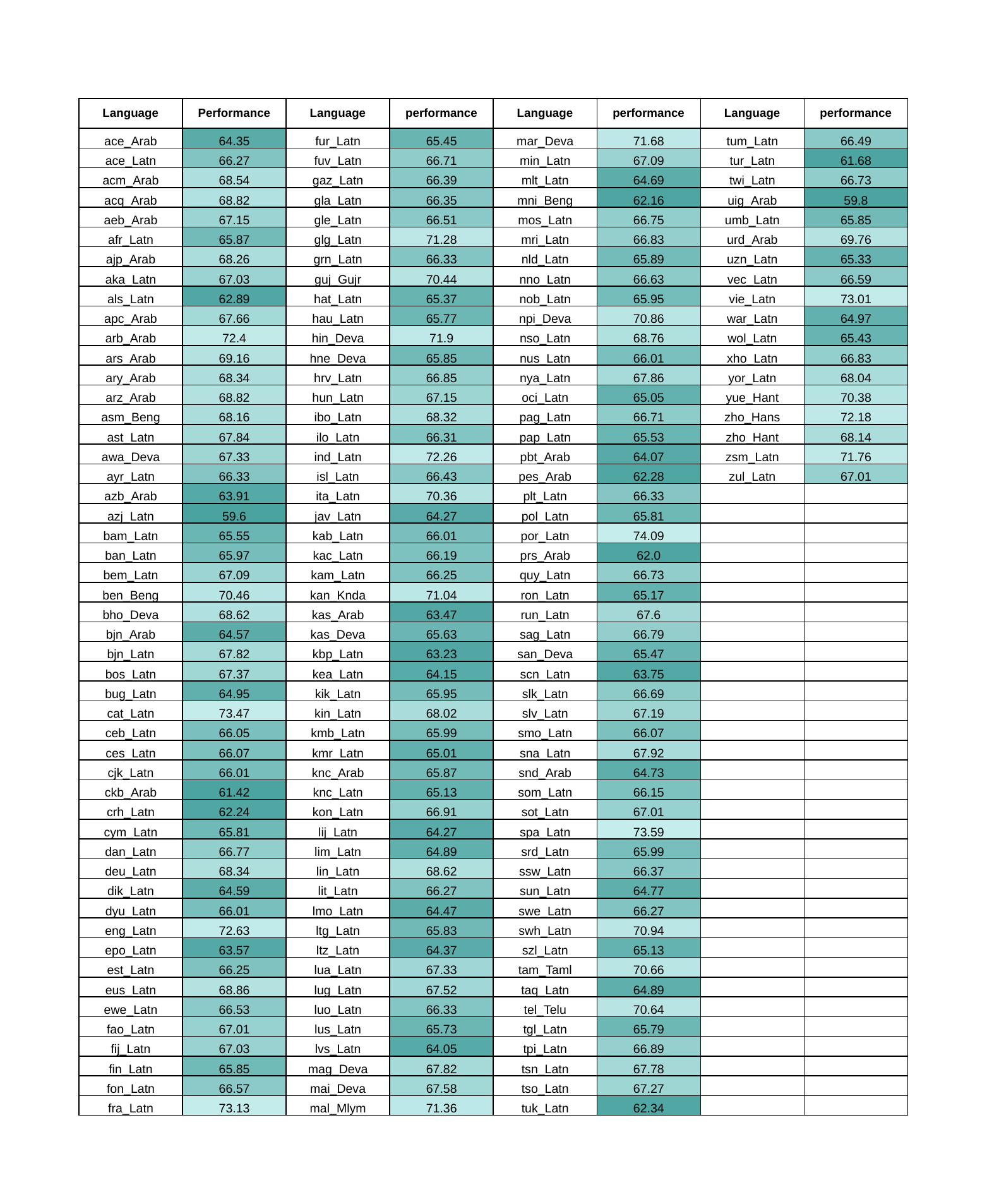}}
    \caption{The accuracy score of BLOOMz model on B-NLI task across 168 languages.}
    \label{fig:bloomz-xnli-detailed-results}
\end{figure}

\begin{figure}[H]
    \centering
    \scalebox{1}{
    \includegraphics[width=\linewidth]{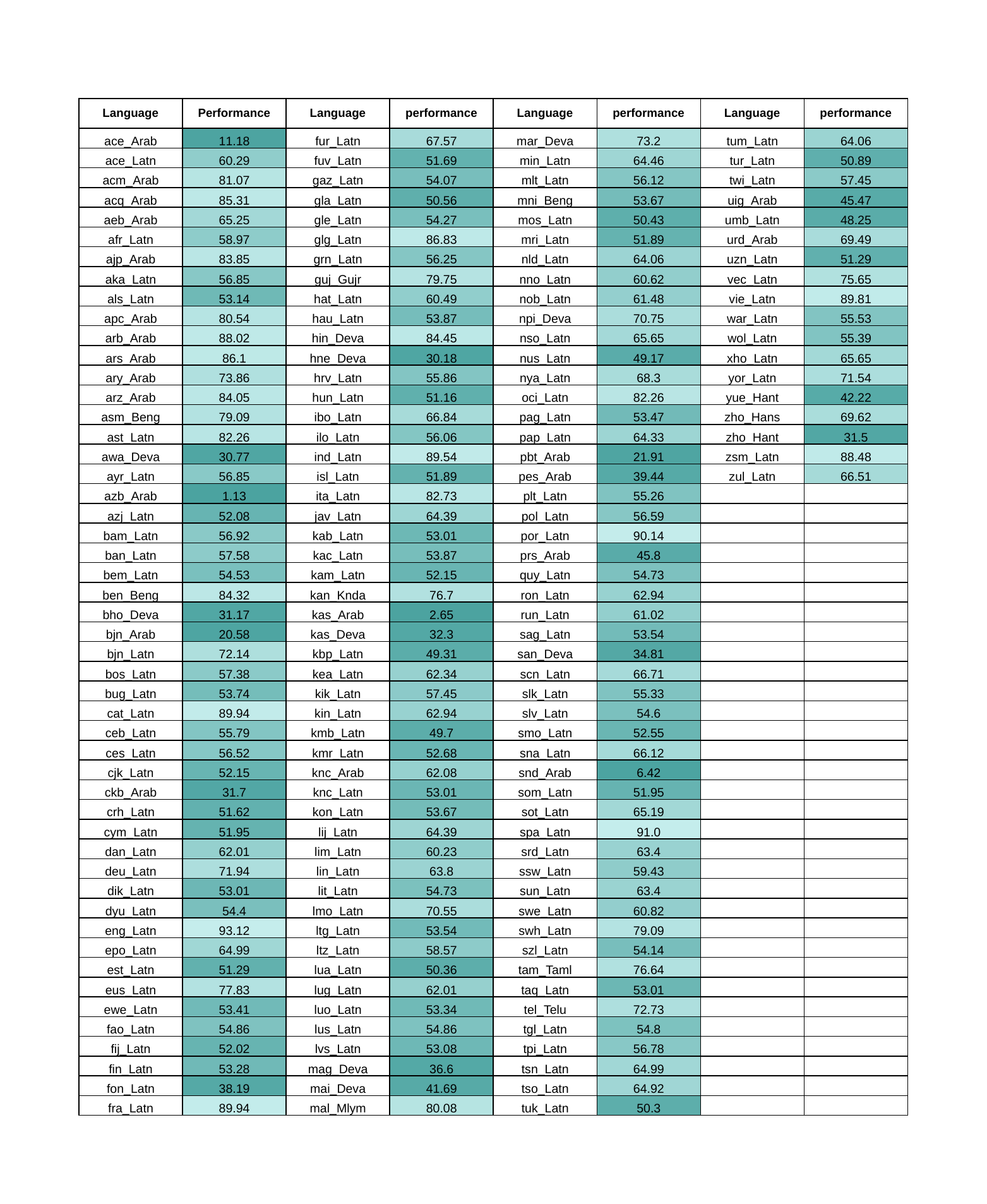}}
    \caption{The accuracy score of BLOOMz model on XStoryCloze task across 168 languages.}
    \label{fig:bloomz-story-detailed-results}
\end{figure}

\begin{figure}[H]
    \centering
    \scalebox{1}{
    \includegraphics[width=\linewidth]{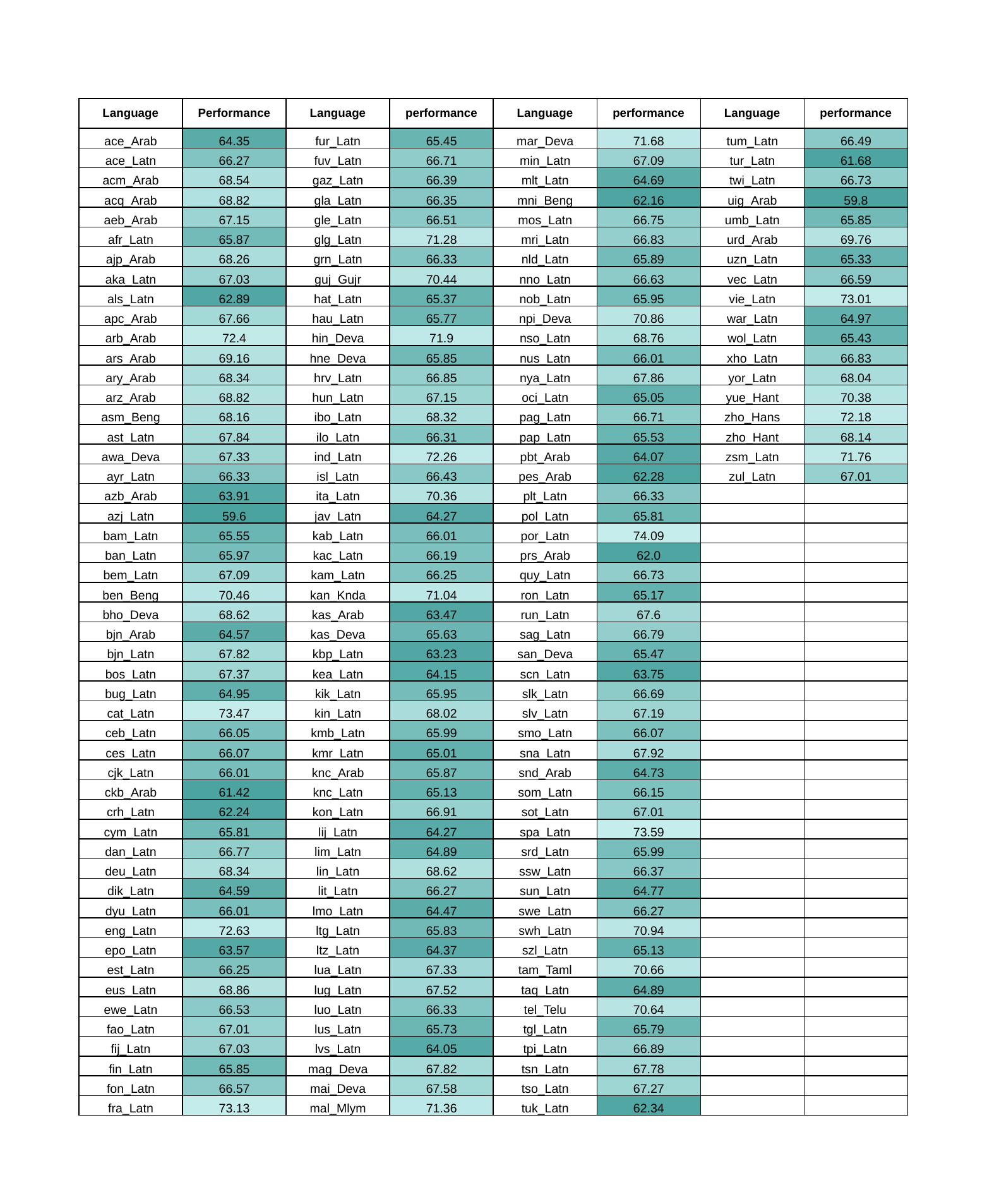}}
    \caption{The accuracy score of BLOOMz model on XCOPA task across 168 languages.}
    \label{fig:bloomz-copa-detailed-results}
\end{figure}

\begin{figure}[H]
    \centering
    \scalebox{1}{
    \includegraphics[width=\linewidth]{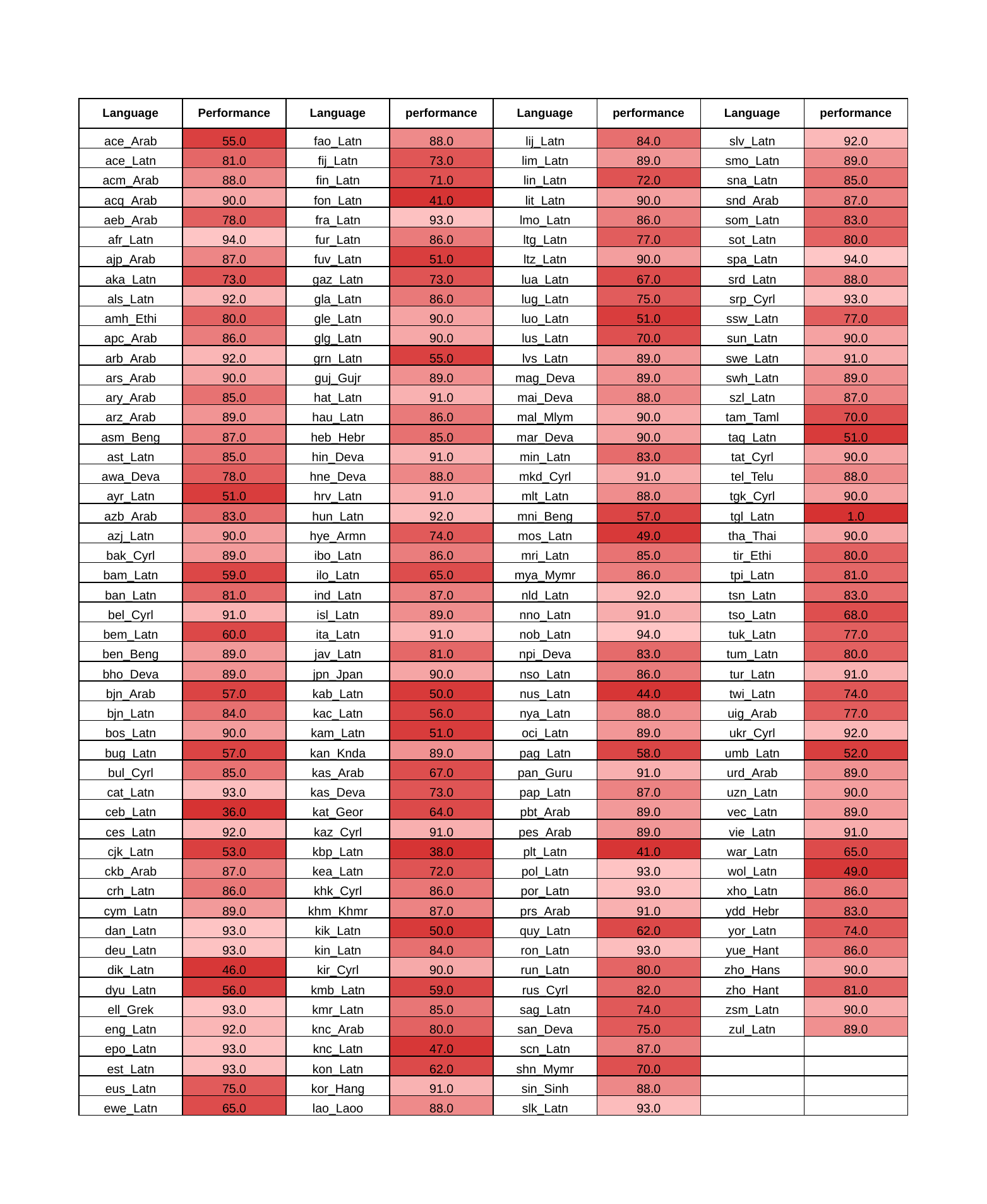}}
    \caption{The accuracy score of AYA on XStoryCloze task across 198 languages.}
    \label{fig:aya-story-detailed-results}
\end{figure}

\begin{figure}[H]
    \centering
    \scalebox{1}{
    \includegraphics[width=\linewidth]{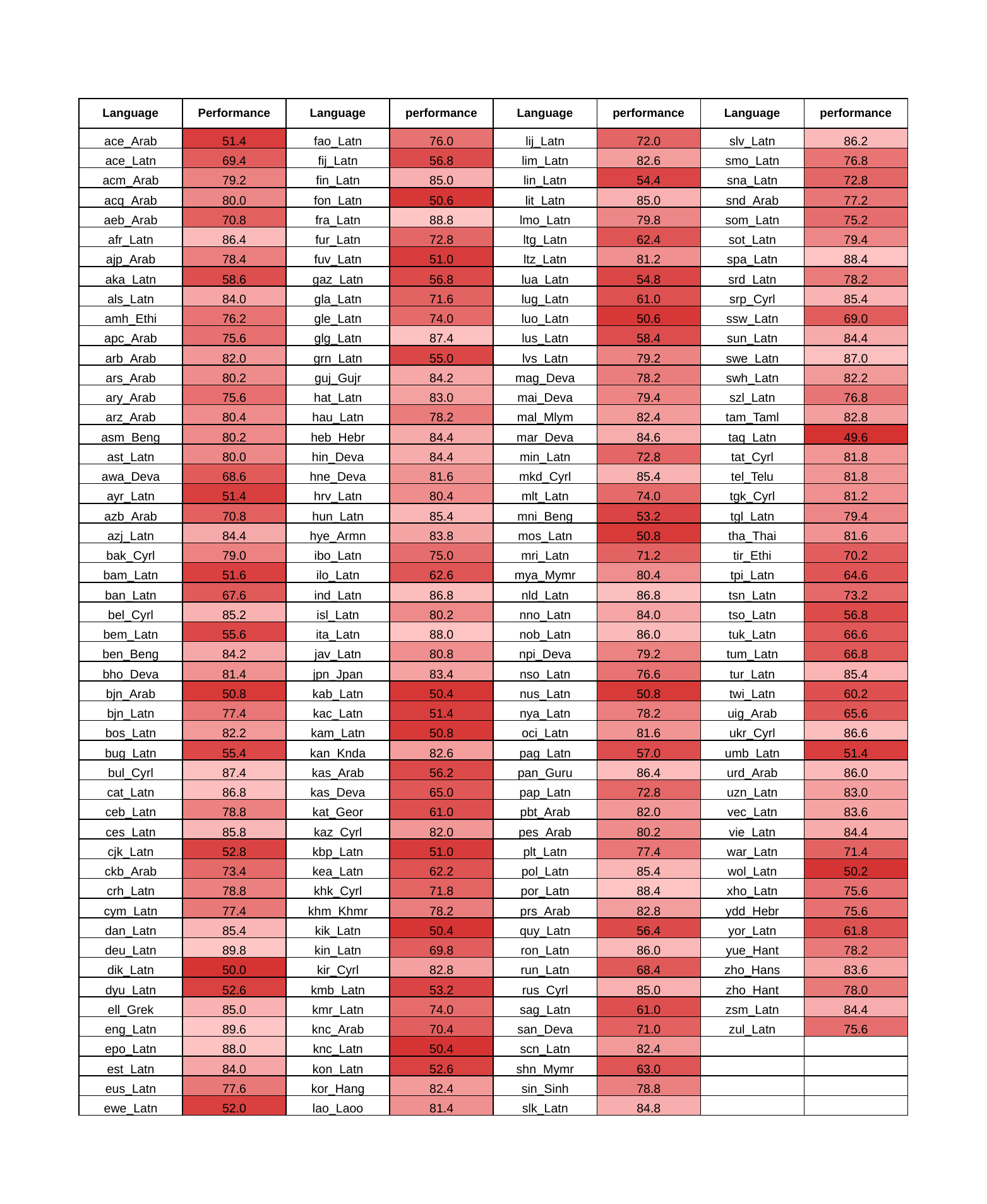}}
    \caption{The accuracy score of AYA on XCOPA task across 198 languages.}
    \label{fig:bloomz-copa-detailed-results}
\end{figure}

\begin{figure}[H]
    \centering
    \scalebox{1}{
    \includegraphics[width=\linewidth]{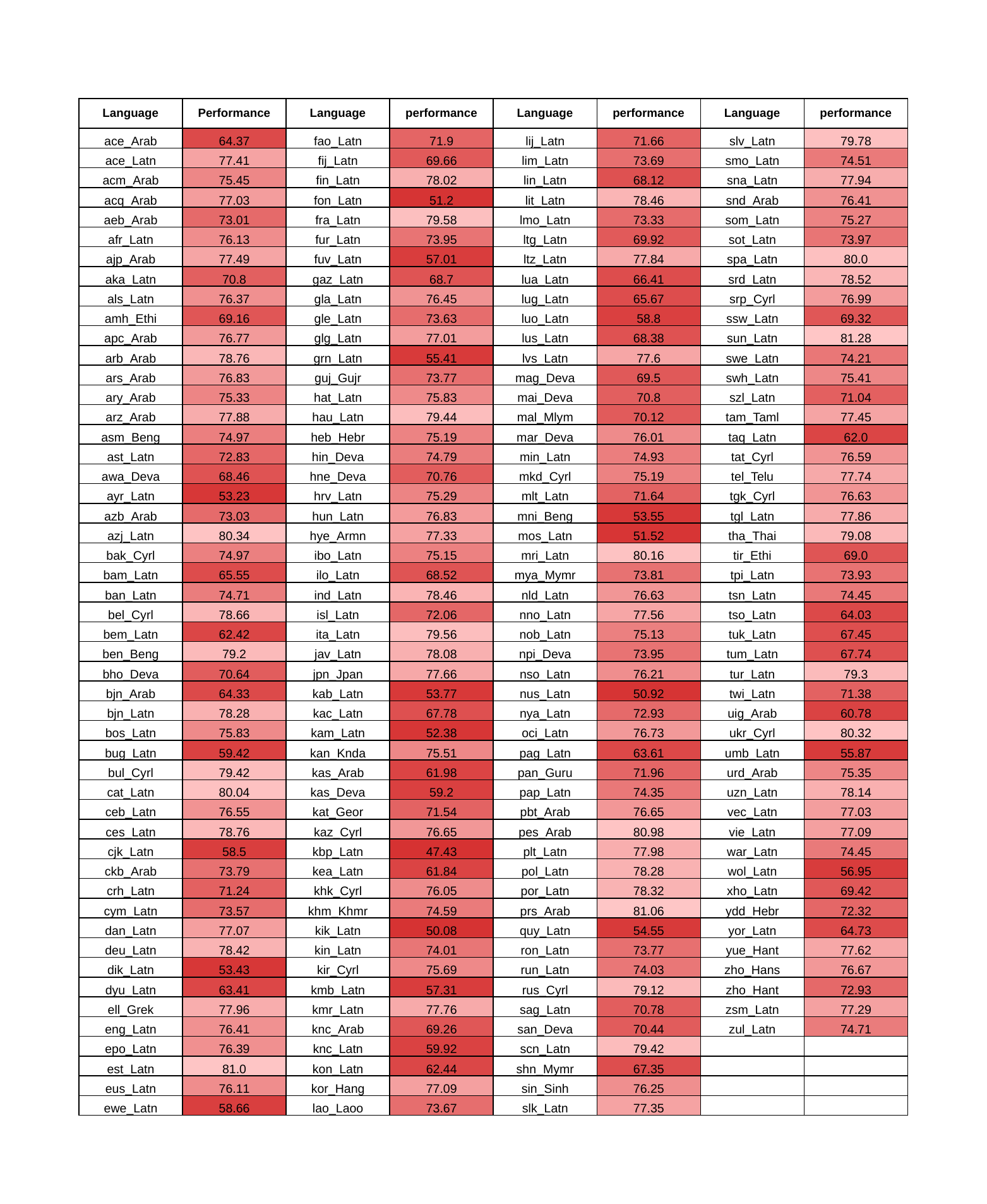}}
    \caption{The accuracy score of AYA on B-NLI task across 198 languages.}
    \label{fig:bloomz-copa-detailed-results}
\end{figure}

\begin{figure}[H]
    \centering
    \scalebox{1}{
    \includegraphics[width=\linewidth]{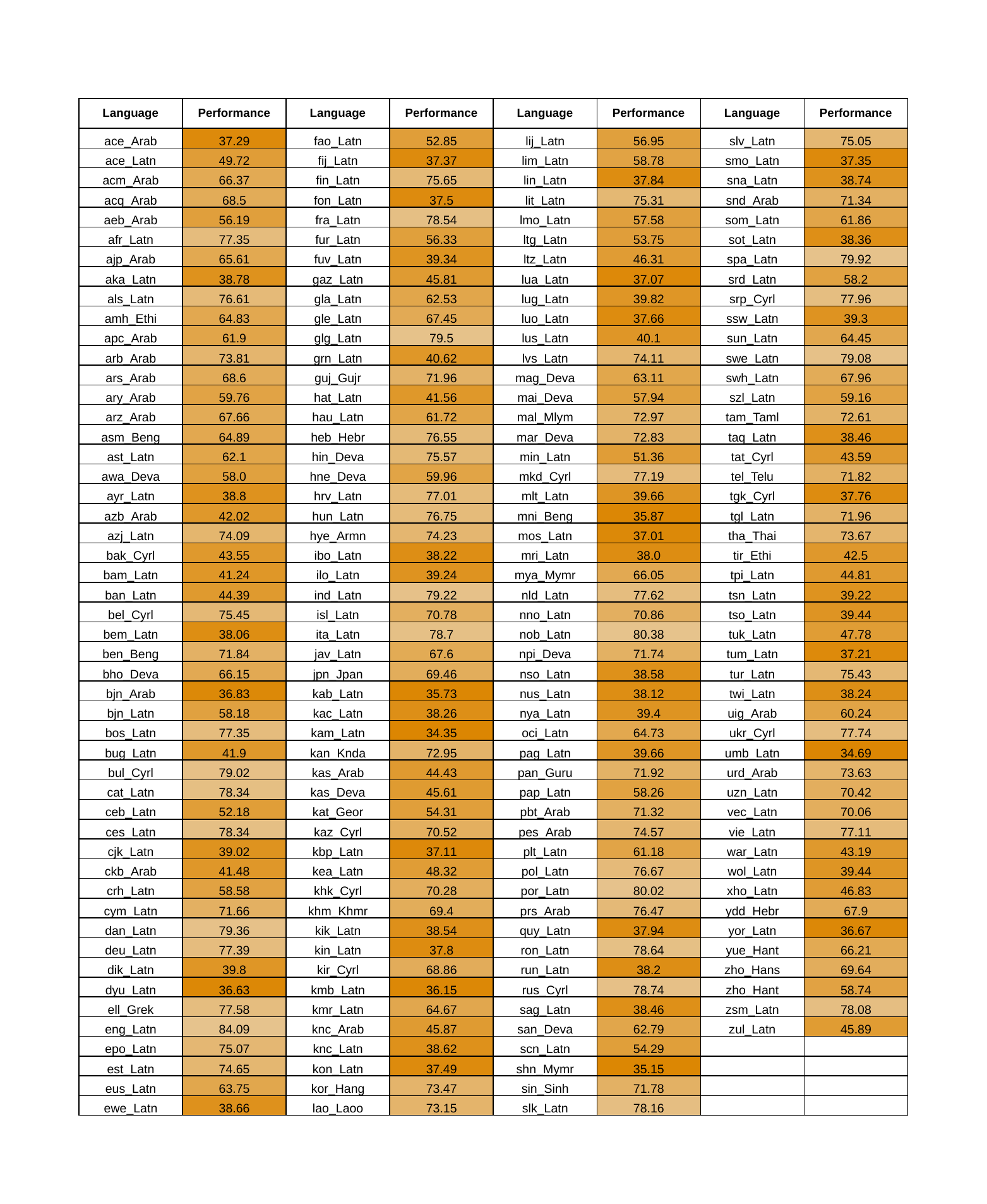}}
    \caption{The accuracy score of XLM-R base on XNLI task across 196 languages.}
    \label{fig:bloomz-copa-detailed-results}
\end{figure}

\begin{figure}[H]
    \centering
    \scalebox{1}{
    \includegraphics[width=\linewidth]{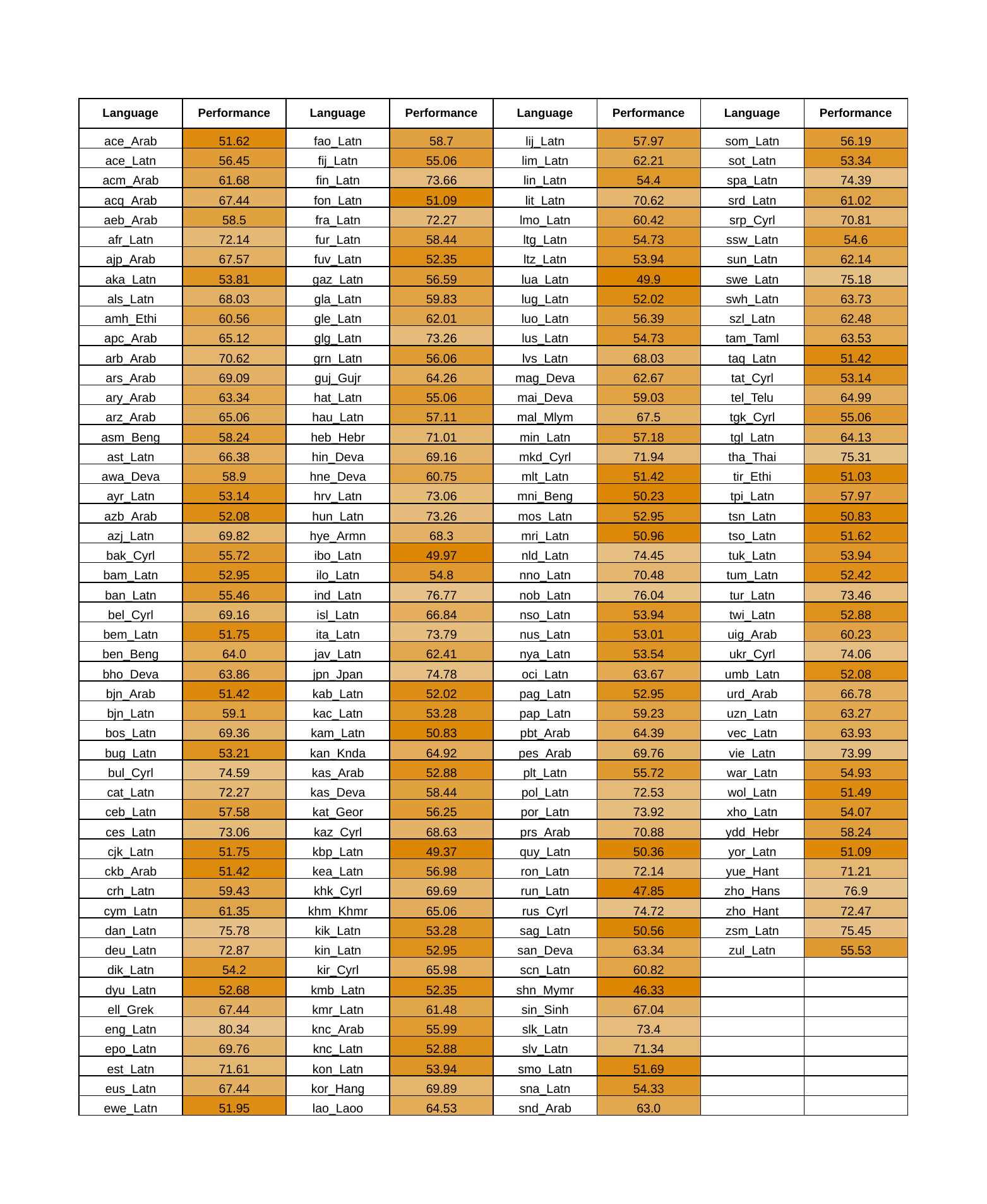}}
    \caption{The accuracy score of XLM-R base on XStoryCloze task across 196 languages.}
    \label{fig:aya-story-detailed-results}
\end{figure}

\begin{figure}[H]
    \centering
    \scalebox{1}{
    \includegraphics[width=\linewidth]{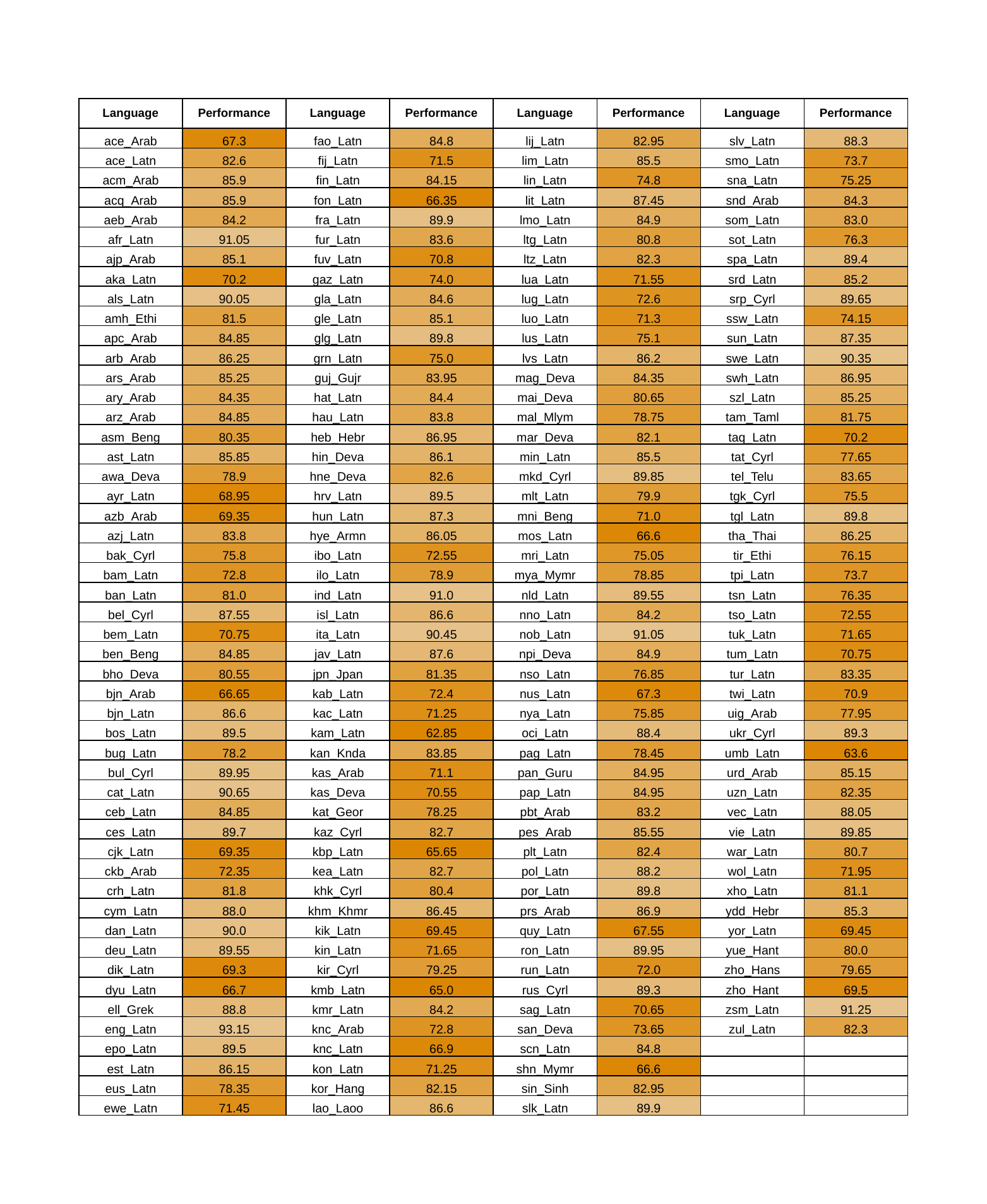}}
    \caption{The accuracy score of XLM-R base on PAWS-X task across 196 languages.}
    \label{fig:bloomz-copa-detailed-results}
\end{figure}

\section{ChrF++ scores per language}\label{sec:chrf_scores}
\begin{figure}[H]
    \centering
    \scalebox{1}{
    \includegraphics[width=\linewidth]{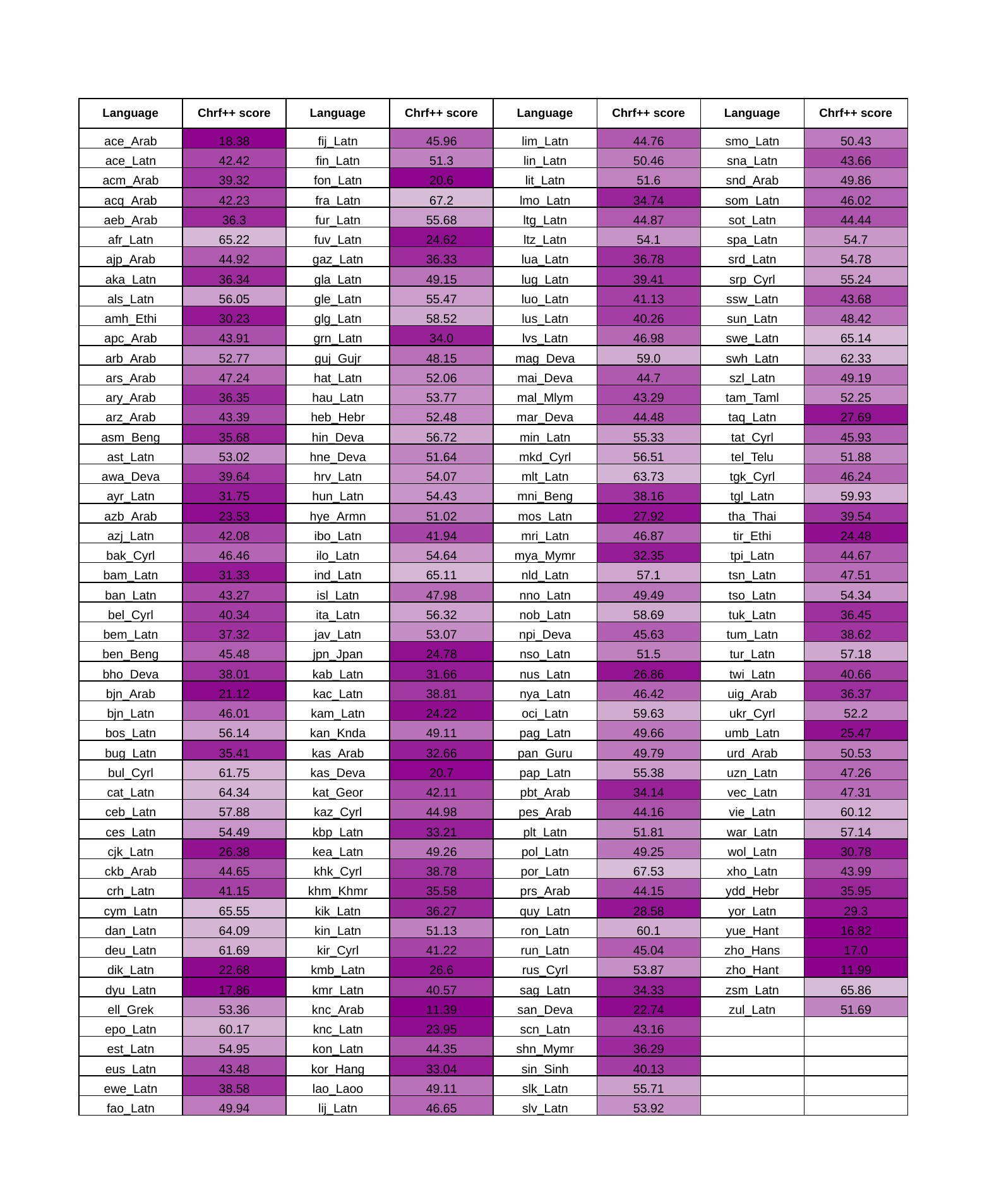}}
    \caption{Chrf++ scores for the selected languages.}
    \label{fig:bloomz-copa-detailed-results}
\end{figure}

\section{Language coverage}\label{sec:languagcoverage}
\begin{table}[H]
    \centering   
    \setlength{\tabcolsep}{6.5pt}
    \scalebox{0.8}{
    \begin{tabular}{|l | r | c  |}
    \toprule
            &Category & Languages \\
    \midrule

     \multirow{30}{*}{\begin{turn}{90}\textbf{XLM-R}\end{turn}} & High  &   arb\_Arab ,
   bul\_Cyrl  ,
   dan\_Latn  ,
   deu\_Latn  ,
   ell\_Grek  , 
   eng\_Latn  ,
   fin\_Latn  ,\\&&
   fra\_Latn  ,
   heb\_Hebr  ,
   hun\_Latn  ,
   ind\_Latn  ,
   ita\_Latn  ,
   jpn\_Jpan  ,
   kor\_Hang  ,\\&&
   nld\_Latn  ,
   nob\_Latn  ,
   pes\_Arab  ,
   pol\_Latn  ,
   por\_Latn  ,
   ron\_Latn  ,
   rus\_Cyrl  ,\\&&
   spa\_Latn  ,
   tha\_Thai  ,
   ukr\_Cyrl  ,
   vie\_Latn  ,
   zho\_Hans  \\
     \cline{2-3}
     &Mid &    als\_Latn  ,
   azj\_Latn  ,
   bel\_Cyrl  ,
   ben\_Beng  ,
   cat\_Latn  ,
   ces\_Latn  ,
   est\_Latn  ,\\&&
   glg\_Latn  ,
   hin\_Deva  ,
   hrv\_Latn  ,
   hye\_Armn  ,
   isl\_Latn  ,
   kan\_Knda  ,
   kat\_Geor  ,\\&&
   kaz\_Cyrl  ,
   khk\_Cyrl  ,
   lit\_Latn  ,
   lvs\_Latn  ,
   mal\_Mlym  ,
   mar\_Deva  ,
   mkd\_Cyrl  ,\\&&
   npi\_Deva  ,
   sin\_Sinh  ,
   slk\_Latn  ,
   slv\_Latn  ,
   srp\_Cyrl  ,
   swe\_Latn  ,
   tam\_Taml  ,\\&&
   tel\_Telu  ,
   tgl\_Latn  ,
   tur\_Latn  ,
   urd\_Arab  ,
   zho\_Hant  ,
   zsm\_Latn   \\
     \cline{2-3}
     &Low &   afr\_Latn  ,
   amh\_Ethi  ,
   asm\_Beng  ,
   bos\_Latn  ,
   ckb\_Arab  ,
   cym\_Latn  ,
   epo\_Latn  ,\\&&
   eus\_Latn  ,
   gaz\_Latn  ,
   gla\_Latn  ,
   gle\_Latn  ,
   guj\_Gujr  ,
   hau\_Latn  ,
   jav\_Latn  ,\\&&
   khm\_Khmr  ,
   kir\_Cyrl  ,
   lao\_Laoo  ,
   mya\_Mymr  ,
   pan\_Guru  ,
   pbt\_Arab  ,
   plt\_Latn  ,\\&&
   san\_Deva  ,
   snd\_Arab  ,
   som\_Latn  ,
   sun\_Latn  ,
   swh\_Latn  ,
   uig\_Arab  ,
   uzn\_Latn  ,\\&&
   xho\_Latn  ,
   ydd\_Hebr    \\
     \cline{2-3}
     &Unseen &   ace\_Arab  ,
   ace\_Latn  ,
   acm\_Arab  ,
   acq\_Arab  ,
   aeb\_Arab  ,
   ajp\_Arab  ,
   aka\_Latn  ,\\&&
   apc\_Arab  ,
   ars\_Arab  ,
   ary\_Arab  ,
   arz\_Arab  ,
   ast\_Latn  ,
   awa\_Deva  ,
   ayr\_Latn  ,\\&&
   azb\_Arab  ,
   bak\_Cyrl  ,
   bam\_Latn  ,
   ban\_Latn  ,
   bem\_Latn  ,
   bho\_Deva  ,
   bjn\_Arab  ,\\&&
   bjn\_Latn  ,
   bug\_Latn  ,
   ceb\_Latn  ,
   cjk\_Latn  ,
   crh\_Latn  ,
   dik\_Latn  ,
   dyu\_Latn  ,\\&&
   ewe\_Latn  ,
   fao\_Latn  ,
   fij\_Latn  ,
   fon\_Latn  ,
   fur\_Latn  ,
   fuv\_Latn  ,
   grn\_Latn  ,\\&&
   hat\_Latn  ,
   hne\_Deva  ,
   ibo\_Latn  ,
   ilo\_Latn  ,
   kab\_Latn  ,
   kac\_Latn  ,
   kam\_Latn  ,\\&&
   kas\_Arab  ,
   kas\_Deva  ,
   kbp\_Latn  ,
   kea\_Latn  ,
   kik\_Latn  ,
   kin\_Latn  ,
   kmb\_Latn  ,\\&&
   kmr\_Latn  ,
   knc\_Arab  ,
   knc\_Latn  ,
   kon\_Latn  ,
   lij\_Latn  ,
   lim\_Latn  ,
   lin\_Latn  ,\\&&
   lmo\_Latn  ,
   ltg\_Latn  ,
   ltz\_Latn  ,
   lua\_Latn  ,
   lug\_Latn  ,
   luo\_Latn  ,
   lus\_Latn  ,\\&&
   mag\_Deva  ,
   mai\_Deva  ,
   min\_Latn  ,
   mlt\_Latn  ,
   mni\_Beng  ,
   mos\_Latn  ,
   mri\_Latn  ,\\&&
   nno\_Latn  ,
   nso\_Latn  ,
   nus\_Latn  ,
   nya\_Latn  ,
   oci\_Latn  ,
   pag\_Latn  ,
   pap\_Latn  ,\\&&
   prs\_Arab  ,
   quy\_Latn  ,
   run\_Latn  ,
   sag\_Latn  ,
   scn\_Latn  ,
   shn\_Mymr  ,
   smo\_Latn  ,\\&&
   sna\_Latn  ,
   sot\_Latn  ,
   srd\_Latn  ,
   ssw\_Latn  ,
   szl\_Latn  ,
   taq\_Latn  ,
   tat\_Cyrl  ,\\&&
   tgk\_Cyrl  ,
   tir\_Ethi  ,
   tpi\_Latn  ,
   tsn\_Latn  ,
   tso\_Latn  ,
   tuk\_Latn  ,
   tum\_Latn  ,\\&&
   twi\_Latn  ,
   umb\_Latn  ,
   vec\_Latn  ,
   war\_Latn  ,
   wol\_Latn  ,
   yor\_Latn  ,
   yue\_Hant  ,\\&&
   zul\_Latn  \\
   \midrule
   \midrule
   \multirow{28}{*}{\begin{turn}{90}\textbf{BLOOMz}\end{turn}} & High  &  arb\_Arab  ,
   cat\_Latn  ,
   eng\_Latn  ,
   fra\_Latn  ,
   ind\_Latn  ,
   por\_Latn  ,
   spa\_Latn  ,\\&&
   vie\_Latn  ,
   zho\_Hans  \\

   \cline{2-3}
     &Mid &    ben\_Beng  ,
   eus\_Latn  ,
   hin\_Deva  ,
   mal\_Mlym  ,
   tam\_Taml  ,
   urd\_Arab  ,
   zho\_Hant'\\
    \cline{2-3}
     &Low &     aka\_Latn  ,
   asm\_Beng  ,
   bam\_Latn  ,
   bho\_Deva  ,
   fon\_Latn  ,
   guj\_Gujr  ,
   ibo\_Latn  ,\\&&
   kan\_Knda  ,
   kik\_Latn  ,
   kin\_Latn  ,
   lin\_Latn  ,
   mar\_Deva  ,
   npi\_Deva  ,
   nso\_Latn  ,\\&&
   sot\_Latn  ,
   swh\_Latn  ,
   tel\_Telu  ,
   wol\_Latn  ,
   xho\_Latn  ,
   yor\_Latn  ,
   zul\_Latn  \\
   \cline{2-3}
     &Unseen &       ace\_Arab  ,
   ace\_Latn  ,
   acm\_Arab  ,
   acq\_Arab  ,
   aeb\_Arab  ,
   afr\_Latn  ,
   ajp\_Arab  ,\\&&
   apc\_Arab  ,
   ars\_Arab  ,
   ary\_Arab  ,
   arz\_Arab  ,
   ast\_Latn  ,
   awa\_Deva  ,
   ayr\_Latn  ,\\&&
   azb\_Arab  ,
   azj\_Latn  ,
   ban\_Latn  ,
   bem\_Latn  ,
   bjn\_Arab  ,
   bjn\_Latn  ,
   bos\_Latn  ,\\&&
   bug\_Latn  ,
   ceb\_Latn  ,
   ces\_Latn  ,
   cjk\_Latn  ,
   ckb\_Arab  ,
   crh\_Latn  ,
   cym\_Latn  ,\\&&
   dan\_Latn  ,
   deu\_Latn  ,
   dik\_Latn  ,
   dyu\_Latn  ,
   epo\_Latn  ,
   est\_Latn  ,
   ewe\_Latn  ,\\&&
   fao\_Latn  ,
   fij\_Latn  ,
   fin\_Latn  ,
   fur\_Latn  ,
   fuv\_Latn  ,
   gla\_Latn  ,
   gle\_Latn  ,\\&&
   glg\_Latn  ,
   grn\_Latn  ,
   hat\_Latn  ,
   hau\_Latn  ,
   hne\_Deva  ,
   hrv\_Latn  ,
   hun\_Latn  ,\\&&
   ilo\_Latn  ,
   isl\_Latn  ,
   ita\_Latn  ,
   jav\_Latn  ,
   kab\_Latn  ,
   kac\_Latn  ,
   kam\_Latn  ,\\&&
   kas\_Arab  ,
   kas\_Deva  ,
   knc\_Arab  ,
   knc\_Latn  ,
   kbp\_Latn  ,
   kea\_Latn  ,
   kmb\_Latn  ,\\&&
   kmr\_Latn  ,
   kon\_Latn  ,
   lij\_Latn  ,
   lim\_Latn  ,
   lit\_Latn  ,
   lmo\_Latn  ,
   ltg\_Latn  ,\\&&
   ltz\_Latn  ,
   lua\_Latn  ,
   lug\_Latn  ,
   luo\_Latn  ,
   lus\_Latn  ,
   lvs\_Latn  ,
   mag\_Deva  ,\\&&
   mai\_Deva  ,
   min\_Latn  ,
   plt\_Latn  ,
   mlt\_Latn  ,
   mni\_Beng  ,
   mos\_Latn  ,
   mri\_Latn  ,\\&&
   nld\_Latn  ,
   nno\_Latn  ,
   nob\_Latn  ,
   nus\_Latn  ,
   nya\_Latn  ,
   oci\_Latn  ,
   gaz\_Latn  ,\\&&
   pag\_Latn  ,
   pap\_Latn  ,
   pes\_Arab  ,
   pol\_Latn  ,
   prs\_Arab  ,
   pbt\_Arab  ,
   quy\_Latn  ,\\&&
   ron\_Latn  ,
   run\_Latn  ,
   sag\_Latn  ,
   san\_Deva  ,
   scn\_Latn  ,
   slk\_Latn  ,
   slv\_Latn  ,\\&&
   smo\_Latn  ,
   sna\_Latn  ,
   snd\_Arab  ,
   som\_Latn  ,
   als\_Latn  ,
   srd\_Latn  ,
   ssw\_Latn  ,\\&&
   sun\_Latn  ,
   swe\_Latn  ,
   szl\_Latn  ,
   tgl\_Latn  ,
   taq\_Latn  ,
   tpi\_Latn  ,
   tsn\_Latn  ,\\&&
   tso\_Latn  ,
   tuk\_Latn  ,
   tum\_Latn  ,
   tur\_Latn  ,
   twi\_Latn  ,
   uig\_Arab  ,
   umb\_Latn  ,\\&&
   uzn\_Latn  ,
   vec\_Latn  ,
   war\_Latn  ,
   yue\_Hant  ,
   zsm\_Latn  \\
    \bottomrule
    \end{tabular} }
    \caption{The languages covered during pretraining of each of the MLMs categorized by the amount of data that was seen for them during pretraining.}
    \label{tab:languages-list}
\end{table}

\begin{table}[H]
    \centering   
    \setlength{\tabcolsep}{6.5pt}
    \scalebox{0.8}{
    \begin{tabular}{|l | r | c  |}
    \toprule
            &Category & Languages \\
    \midrule

     \multirow{30}{*}{\begin{turn}{90}\textbf{AYA}\end{turn}} & High  &   hye\_Armn ,
    kan\_Knda , 
    tur\_Latn ,
    ita\_Latn , 
    nld\_Latn , 
    pol\_Latn ,
    por\_Latn , \\&&
    isl\_Latn ,
    fra\_Latn ,
    deu\_Latn ,
    spa\_Latn , 
    rus\_Cyrl ,
    eng\_Latn \\
     \cline{2-3}
     &Mid &    est\_Latn ,
    ben\_Beng ,
    mar\_Deva ,
    slv\_Latn ,
    lit\_Latn ,
    heb\_Hebr ,
    zsm\_Latn , \\&&
    cat\_Latn , 
    tha\_Thai , 
    kor\_Hang ,
    slk\_Latn , 
    hin\_Deva ,
    bul\_Cyrl ,
    nob\_Latn , \\&&
    fin\_Latn ,
    dan\_Latn ,
    hun\_Latn ,
    ukr\_Cyrl ,
    ell\_Grek ,
    ron\_Latn ,
    swe\_Latn , \\&&
    arb\_Arab ,
    pes\_Arab ,
    zho\_Hans ,
    ces\_Latn  \\
     \cline{2-3}
     &Low &   hat\_Latn ,
    kor\_Hang ,
    xho\_Latn ,
    ibo\_Latn ,
    lao\_Laoo ,
    mri\_Latn ,
    smo\_Latn ,\\&&
    ckb\_Arab ,
    amh\_Ethi ,
    nya\_Latn ,
    hau\_Latn ,
    plt\_Latn ,
    pbt\_Arab ,
    gla\_Latn ,\\&&
    sun\_Latn ,
    jpn\_Jpan ,
    sot\_Latn ,
    ceb\_Latn ,
    pan\_Guru ,
    gle\_Latn ,
    kir\_Cyrl , \\&&
    epo\_Latn , 
    sin\_Sinh , 
    guj\_Gujr ,
    yor\_Latn ,
    tgk\_Cyrl ,
    snd\_Arab ,
    mya\_Mymr , \\&&
    kaz\_Cyrl ,
    khm\_Khmr ,
    som\_Latn ,
    swh\_Latn , 
    ydd\_Hebr ,
    uzn\_Latn ,
    hun\_Latn , \\&&
    mlt\_Latn , 
    eus\_Latn ,
    bel\_Cyrl ,
    kat\_Geor ,
    mkd\_Cyrl ,
    mal\_Mlym ,
    khk\_Cyrl , \\&&
    tha\_Thai ,
    afr\_Latn ,
    ukr\_Cyrl ,
    ltz\_Latn ,
    tel\_Telu ,
    urd\_Arab ,
    lit\_Latn , \\&&
    npi\_Deva ,
    srp\_Cyrl ,
    tam\_Taml ,
    cym\_Latn ,
    als\_Latn ,
    glg\_Latn ,
    azj\_Latn ,\\&&
    lvs\_Latn  \\
     \cline{2-3}
     &Unseen &   ace\_Arab ,
    ace\_Latn ,
    acm\_Arab ,
    acq\_Arab ,
    aeb\_Arab ,
    ajp\_Arab ,
    aka\_Latn , \\&&
    apc\_Arab ,
    ars\_Arab , 
    ary\_Arab ,
    arz\_Arab ,
    asm\_Beng ,
    ast\_Latn ,
    awa\_Deva , \\&&
    ayr\_Latn ,
    azb\_Arab ,
    bak\_Cyrl ,
    bam\_Latn ,
    ban\_Latn ,
    bem\_Latn ,
    bho\_Deva , \\&&
    bjn\_Arab , 
    bjn\_Latn ,
    bos\_Latn ,
    bug\_Latn ,
    cjk\_Latn ,
    crh\_Latn ,
    dik\_Latn , \\&&
    dyu\_Latn ,
    ewe\_Latn ,
    fao\_Latn , 
    fij\_Latn ,
    fon\_Latn ,
    fur\_Latn ,
    fuv\_Latn , \\&&
    grn\_Latn ,
    hne\_Deva , 
    hrv\_Latn ,
    ilo\_Latn ,
    kab\_Latn ,
    kac\_Latn ,
    kam\_Latn , \\&&
    kas\_Arab ,
    kas\_Deva ,
    knc\_Arab ,
    knc\_Latn ,
    kbp\_Latn ,
    kea\_Latn ,
    kik\_Latn , \\&&
    kin\_Latn ,
    kmb\_Latn ,
    kmr\_Latn ,
    kon\_Latn ,
    lij\_Latn ,
    lim\_Latn ,
    lin\_Latn , \\&&
    lmo\_Latn , 
    ltg\_Latn ,
    lua\_Latn ,
    lug\_Latn ,
    luo\_Latn ,
    lus\_Latn ,
    mag\_Deva ,\\&&
    mai\_Deva ,
    min\_Latn ,
    mni\_Beng ,
    mos\_Latn ,
    nno\_Latn , 
    nso\_Latn ,
    nus\_Latn ,\\&&
    oci\_Latn ,
    gaz\_Latn ,
    pag\_Latn ,
    pap\_Latn ,
    prs\_Arab ,
    quy\_Latn ,
    run\_Latn , \\&&
    sag\_Latn ,
    san\_Deva ,
    scn\_Latn ,
    shn\_Mymr ,
    srd\_Latn ,
    ssw\_Latn ,
    szl\_Latn , \\&&
    tat\_Cyrl  ,
    tgl\_Latn ,
    tir\_Ethi ,
    taq\_Latn ,
    tpi\_Latn ,
    tsn\_Latn ,
    tso\_Latn , \\&&
    tuk\_Latn ,
    tum\_Latn , 
    twi\_Latn ,
    tzm\_Tfng ,
    uig\_Arab ,
    umb\_Latn ,
    vec\_Latn , \\&&
    war\_Latn ,
    wol\_Latn ,
    yue\_Hant ,
    zho\_Hant ,
    dzo\_Tibt  \\
    \bottomrule
    \end{tabular} }
    \caption{The languages covered during AYA's pretraining categorized by the amount of data that was seen during pretraining.}
    \label{tab:languages-list}
\end{table}

\section{Full Results for XLM-R base}

\begin{table*}[!ht]
    \centering   
    \setlength{\tabcolsep}{2pt}
    \scalebox{0.59}{
    \begin{tabular}{l  c c c c c c c c c c  c c c c c c c c c c c cc cc }
    \toprule
                 & ar & bg & de & el &  es & et &eu & fr& hi & ht& id& it& ja& ko & my& qu& ru& sw &  ta& te& th& tr& ur& vi& zh  \\
                   \midrule 
        & \multicolumn{25}{c}{\textbf{XLM-R base}} \\
        \midrule
        
          XStoryCloze & 68/70 & - & - & - & 74/74 & - & 67/65 & - & 69/69 & - & 76/76 & - & - & - & 63/65 & - & 75/74 & 64/63 &  -& 68/65 & -& -&-& - & 77/77 \\
          XNLI & 71/73 & 78/79 & 75/77 & 75/77 & 79/75 & - & - & 78/78 & 65/67 & - & - & -& - & - & - & -& 75/78 & 70/74 &  -& - & 71/73&73/75& 65/73& 74/77& 73/69 \\
          PAWS-X & - & - & 87/89& - & 88/89 & -& - & 89/89 & - &  -& - & -& 77/81 & 76/82 & - & - & - & - &  -& - & - & - &-& -& 82/79 \\

\bottomrule
\end{tabular}}
    \caption{{The performance in ($\%$) accuracy when evaluating the XLM-R base on the human translated (original) datasets/our machine translated datasets.}} 
    \label{tab:translation-quality}
\end{table*}










\end{document}